\documentclass[5p, twocolumn]{elsarticle}
\usepackage{array}
\usepackage{multicol}
\usepackage{multirow}
\usepackage{amsmath}
\usepackage{lineno,hyperref}
\usepackage{subfig}
\usepackage{threeparttable}
\journal{Biomedical Signal Processing and Control}

\begin{document}

\begin{frontmatter}

\title{A Recurrent Convolutional Neural Network Approach\\for Sensorless Force Estimation in Robotic Surgery}


\author[upc_address,fhhi_address]{Arturo~Marban\corref{corresponding_author}}
\ead{arturo.marban@upc.edu}

\author[fhhi_address]{Vignesh~Srinivasan}
\ead{vignesh.srinivasan@hhi.fraunhofer.de}

\author[fhhi_address]{Wojciech~Samek}
\ead{wojciech.samek@hhi.fraunhofer.de}

\author[upc_address]{Josep {Fern\'{a}ndez}}
\ead{josep.fernandez@upc.edu}

\author[upc_address]{Alicia Casals}
\ead{alicia.casals@upc.edu}

\address[upc_address]{Research Centre for Biomedical Engineering (CREB),\\Universitat Polit\`{e}cnica de Catalunya, 08034 Barcelona, Spain}
\address[fhhi_address]{Machine Learning Group, Fraunhofer Heinrich Hertz Institute, 10587 Berlin, Germany}

\cortext[corresponding_author]{Corresponding author}

%
%
\begin{abstract}
Providing force feedback as relevant information in current Robot-Assisted Minimally Invasive Surgery systems constitutes a technological challenge due to the constraints imposed by the surgical environment. In this context, Sensorless Force Estimation techniques represent a potential solution, enabling to sense the interaction forces between the surgical instruments and soft-tissues. Specifically, if visual feedback is available for observing soft-tissues' deformation, this feedback can be used to estimate the forces applied to these tissues. To this end, a force estimation model, based on Convolutional Neural Networks and Long-Short Term Memory networks, is proposed in this work. This model is designed to process both, the spatiotemporal information present in video sequences and the temporal structure of tool data (the surgical tool-tip trajectory and its grasping status).  A series of analyses are carried out to reveal the advantages of the proposal and the challenges that remain for real applications. This research work focuses on two surgical task scenarios, referred to as pushing and pulling tissue. For these two scenarios, different input data modalities and their effect on the force estimation quality are investigated. These input data modalities are tool data, video sequences and a combination of both. The results suggest that the force estimation quality is better when both, the tool data and video sequences, are processed by the neural network model. Moreover, this study reveals the need for a loss function, designed to promote the modeling of smooth and sharp details found in force signals. Finally, the results show that the modeling of forces due to pulling tasks is more challenging than for the simplest pushing actions. 
\end{abstract}

\begin{keyword}
Robotic Surgery\sep Sensorless Force Estimation\sep Convolutional Neural Networks\sep LSTM Networks.
\end{keyword}

\end{frontmatter}


%
%
\section{Introduction}
\label{section:introduction}
Traditional open surgery, characterized by long incisions, has been improved by minimally invasive surgery, which uses long instruments inserted into the body through small incisions. An endoscopic camera provides visual feedback of the target scenario, and two or more surgical instruments allow the surgeon to interact with tissues and organs. Minimally invasive surgery has been extended and enhanced in capabilities by robotic teleoperated systems with a master-slave configuration, resulting in a new procedure known as Robotic Assisted Minimally Invasive Surgery (RAMIS)~\cite{Palep_2009}\cite{Gomes_2011}.

RAMIS provides surgeons with augmented capabilities, such as fine and dexterous movements, proper hand-eye coordination, hand tremor suppression and high-quality visualization of the surgical scenario~\cite{Gomes_2011}. Nonetheless, the integration of force feedback as relevant information in these systems still remains an open problem~\cite{Marban_2014}\cite{Bayle_2014}. Force feedback has proven to be beneficial in teleoperated surgery since it is associated with the control of interaction forces and thus, its use can result in less intraoperative tissue damage produced by the application of excessive forces. Force feedback also helps to improve the proper execution of surgical tasks, such as grasping or suturing, in which the application of excessive or insufficient forces can produce damage or malfunctions. Furthermore, force feedback can provide information of tissue stiffness and shape. Therefore, it can help to detect abnormalities, such as tumors or calcified arteries~\cite{Okamura_2_2011}.

The main difficulty in providing RAMIS systems with force feedback relies on measuring interaction forces between surgical instruments and tissues. This problem can be addressed by two approaches: direct force sensing and sensorless force estimation. In direct force sensing, the measurement of forces is carried out with a sensor located at, or close to, the point of interaction between tool and tissue. Although it represents the most intuitive solution, many constraints, such as biocompatibility, sterilization, miniaturization, and cost~\cite{Okamura_2009}, limit the design of such force sensors. The need of miniaturization has been addressed in different works such as~\cite{DH_Lee_2016}, where a laparoscopic instrument with force sensing capability is described. However, its clinical validation has not been proven yet, since it was only tested in an open platform for surgical robotics research, called Raven-II~\cite{Hannaford_2013}. In contrast, force estimation allows the removal of any electronic device from the instrument in contact with the patient. Therefore, the interaction forces have to be estimated from the available sources of information, which may result in inaccurate measures. Due to the aforesaid reasons, sensorless force estimation represents a potential solution for the practical implementation of force perception systems in RAMIS.

Sensorless force estimation can be implemented through control-based or vision-based approaches. In the control-based approach, interaction forces are estimated using observers and models of the surgical tool, and by processing available information from the motor units (i.e.~angular position/velocity, current consumption, and torque). In this regard, some relevant works are focused on estimating the surgical instrument grasping force, as described in \cite{Yoon_2013} and \cite{Y_Li_2016}. In contrast, the vision-based approach consists in estimating forces mainly from video sequences (monocular or stereo), therefore, in this work it is referred to as Vision-Based Force Sensing (VBFS). In VBFS, the uncertainty of the force estimates is reduced by having access to surgical tool data, such as tool-tip trajectory, its velocity, and grasper status. Although there are fewer works in the literature related to VBFS, if developed properly, it has potential to restore force feedback in robotic surgery. VBFS avoids the need for accurate modeling of the surgical instrument or slave-robot manipulator, as required by most control-based approaches. 

In the next section, deep neural networks are introduced as effective models applied in the processing of video sequences (Section~\ref{section:intro:dnn_for_processing_video_sequences}). Subsequently, the concept of VBFS is defined and different works reported in the literature are described (Section~\ref{section:intro:vision_based_force_sensing}). Finally, the proposed approach for estimating forces in robotic surgery is presented and the contributions of this research work are listed (Section~\ref{section:intro:rcnn_approach}).

%
%
%
%
\subsection{Deep Neural Networks for Processing Video Sequences}
\label{section:intro:dnn_for_processing_video_sequences}

Deep neural networks composed of Convolutional Neural Networks (CNN) and Long-Short Term Memory (LSTM) networks have been investigated in different domains where the input data has a spatiotemporal structure, as in video sequences. The CNN addresses the processing of spatial information, while the LSTM network the processing of temporal information. This neural network architecture has been applied in action recognition with visual attention~\cite{SSharma_2015}, video activity recognition and image captioning~\cite{JDonahue_2015}, video content description~\cite{Venugopalan_2015}, and learning physical interaction through video prediction~\cite{CFinn_2016}, among others. A particular domain of interest is related to the estimation of time-varying signals from video sequences in the context of a regression framework. In this regard,~\cite{Owens_2015} proposed a technique to estimate sound from silent video sequences through a neural network consisting of a CNN and LSTM networks. This neural network was trained using a video dataset, describing interactions of a wooden stick with different objects and materials with added audio recordings. In another application,~\cite{JZhou_2016} developed a technique to estimate continuous pain intensity from video sequences of facial expressions. This technique is based on a CNN with added recurrent connections in its layers.

In the processing of sequences of data with long-term temporal dependencies, LSTM networks have excelled, providing state of the art results in applications such as language modeling and translation, speech synthesis, and analysis of audio and video data~\cite{AGraves_2014}\cite{JChung_2014}\cite{Greff_2016}. In particular, the LSTM network with coupled input-forget gates, suggested by \cite{Greff_2016} as a less computational expensive model than the vanilla LSTM network, was found suitable for the force estimation task, as discussed later in subsection~\ref{subsection:experiments:lstm_optimization}.

%
%
\subsection{Vision-Based Force Sensing}
\label{section:intro:vision_based_force_sensing}
The Vision-Based Force Sensing (VBFS) concept relies on a simple observation, that is, soft bodies made of biological (i.e.~tissue) or artificial (i.e.~silicone) materials deform under an applied load. Therefore, if the deformation of soft bodies (i.e.~biological tissues) is available from visual feedback (i.e.~video sequences), this feedback can be used to estimate the forces applied on these objects,~\cite{VBFS:Wang_2001}\cite{VBFS:Greminger_2003}. VBFS methods are developed to estimate forces in 2D or 3D scenarios. In the first case, a force applied to a soft body results in a deformed contour, while in the second case, it produces a deformed surface.

%
%
Notable works, such as~\cite{VBFS:Greminger_2003} and~\cite{VBFS:Karimirad_2014}, developed the concept of VBFS in 2D scenarios using neural networks. This approach circumvents the explicit modeling of complex mechanical properties attributed to some materials (i.e.~biological cells). In~\cite{VBFS:Greminger_2003}, VBFS is applied to estimate forces in objects that exhibit both linear (a microgripper) and non-linear (a rubber torus) mechanical properties. This method relies on a deformable template matching algorithm to describe the object's contour deformation and a fully-connected neural network that models the object's mechanical properties. The micromanipulation of cells with a spherical shape has been addressed in~\cite{VBFS:Karimirad_2014}. In this work, a method is developed to estimate force during microinjection of zebrafish embryos. This method relies on active contours and conic fitting algorithms to model the cell's contour deformation. Then, a fully-connected neural network learns the non-linear relationship between deformation and force.
%
%

The estimation of interaction forces between tools and tissues becomes more realistic when tissue deformation is processed in 3D space, that is, by taking into account depth information. To this end, a stereo vision system is used to recover such information. Minimally invasive surgical procedures are complex, however, they can be interpreted as the composition of different elementary surgical tasks~\cite{Rosen_2006}. One of such tasks, referred to as pushing tissue (pressing the end of the endoscopic tools against soft-tissue), represents a common practice in minimally invasive surgery~\cite{VBFS:Kennedy_2005}. This surgical task is studied in the context of VBFS due to its simplicity. 

Force estimation techniques that rely on a stereo vision system are reported in \cite{VBFS:Kennedy_2005},~\cite{VBFS:Kim_2012},~\cite{VBFS:Giannarou_2016},~\cite{VBFS:Aviles_2014} and \cite{VBFS:Aviles_3_2016}.
In~\cite{VBFS:Kennedy_2005}, the forces developed in a rubber membrane are studied. Its deformation was recovered by tracking nodal displacements and a finite element method was used to model the mechanical relationship between deformation and force. VBFS applied to neurosurgery was investigated in~\cite{VBFS:Kim_2012}. In this work, soft-tissue surface deformation is computed using a depth map extracted from stereo-endoscopic images. Thereafter, this information is processed by a surface mesh (based on spring-damper models) to render force as output. Another approach in the context of neurosurgery has been investigated in~\cite{VBFS:Giannarou_2016}. The authors of this work developed a method based on quasi-dense stereo correspondence to recover surface deformation from stereo video sequences. Afterward, force is estimated from the surgical tool displacement (which is extracted from the deformation data), using a 2nd order polynomial model. In recent years, models based on neural networks have been investigated. In this regard,~\cite{VBFS:Aviles_2014} proposed a  method consisting in a 3D lattice and a recurrent neural network. The 3D lattice models the complex deformation of soft-tissues. The recurrent neural network was designed to estimate force by processing the information provided by this lattice in addition to the surgical tool motion. A subsequent notable work by the same author is presented in~\cite{VBFS:Aviles_3_2016}. In this work, the recurrent neural network described in~\cite{VBFS:Aviles_2014} is improved by designing a model based on the LSTM network architecture~\cite{Hochreiter_1997}, achieving high accuracy in the estimation of forces (in 3D space).
Monocular force estimation represents a more challenging approach. In this regard,~\cite{VBFS:Noohi_2014} developed a technique to estimate forces from monocular video sequences using a real lamb liver as experimental material. This method relies on a virtual template to model soft-tissue surface deformation, however, it assumes that soft-tissue surface behaves as a smooth function with local deformation. Then, a stress-strain bio-mechanical model defines the relationship between force and penetration depth caused by the surgical tool.

From the literature review a series of conclusions are drawn. First, most of the existing methods recover tissue deformation using a stereo vision system (\cite{VBFS:Kennedy_2005}-\cite{VBFS:Aviles_3_2016}). They rely on a deformation model which is created based on 3D geometries such as a mesh or lattice (i.e.~\cite{VBFS:Kim_2012} and~\cite{VBFS:Aviles_3_2016}), or stereo-correspondences (i.e.~\cite{VBFS:Giannarou_2016}). Second, the estimation of forces has been studied only for pushing tasks. Other surgical tasks that result in complex interactions, such as pulling or grasping tissue, have not been addressed yet. Third, recurrent neural network architectures have been studied in~\cite{VBFS:Aviles_2014} and \cite{VBFS:Aviles_3_2016}, performing a mapping from soft-tissue deformation and tool data to interaction force. From these two works, only~\cite{VBFS:Aviles_3_2016} describes the use of a deep neural network, specifically a LSTM network. Fourth, CNNs, which excel in tasks related to processing spatial information present in images or video sequences (e.g., \cite{Krizhevsky_2012, SSharma_2015, BosTIP18}) have not been explored in the processing of visual information available from RAMIS systems. Fifth, monocular force estimation was only addressed in~\cite{VBFS:Noohi_2014}. Nonetheless, this method relies on feature detection and matching algorithms that are not robust to specularities produced by reflection of light on the tissue surface. Therefore, feature points had to be detected and matched manually during the reported experiments. Furthermore, the force was estimated only for the loading cycle (when the tool is incrementally deforming the tissue, before reaching the peak force), and for one component ($F_z$). Finally, due to the complexity of data acquisition (i.e.~video sequences, tool data and force sensing) in a real surgical scenario, most methods (\cite{VBFS:Kennedy_2005}-\cite{VBFS:Aviles_3_2016}) are implemented and validated on experimental platforms using organs made of artificial tissues (i.e.~silicone). Only~\cite{VBFS:Noohi_2014} describes experiments on a real lamb liver. 

The literature review shows that an approach based on deep neural networks, specifically, CNN and LSTM networks, has not been investigated for VBFS in robotic surgery. Its advantages and downsides will reveal new research directions to design a better force estimation model that learns from data. In particular, transfer learning techniques (i.e.~using a pre-trained CNN on the ImageNet dataset~\cite{Russakovsky_2015}) have not been explored for VBFS in the context of robotic surgery. They can be useful to encode complex phenomena (i.e.~tool-tissue interactions) in a low-dimensional feature vector representation learned from high-dimensional data, such as video sequences. This feature vector representation is easier to model by an LSTM network.

\subsection{Recurrent Convolutional Neural Network Approach}
\label{section:intro:rcnn_approach}

In the present work, a Recurrent Convolutional Neural Network (RCNN) architecture, based on CNN and LSTM networks, is proposed for VBFS in RAMIS. It estimates a 6-dimensional vector of forces and torques (in the 3D space) at every time instant, by processing monocular video sequences and tool data. 

The focus of this research work is on the estimation of interaction forces in two surgical tasks, pushing (pressing the tool against a tissue) and pulling a tissue (which requires grasping). This surgical task decomposition was motivated by the discrete model presented in~\cite{Rosen_2006}. In that work, the complexity of minimally invasive surgical procedures is modeled taking into account a set of fundamental tasks, among them, pushing and pulling a tissue. Moreover, different input data modalities and their effect on the force estimation quality are investigated. These input data modalities are: (i) the tool data represented by the tool-tip trajectory (in 3D space) and its grasping status (opened/closed), (ii) video sequences, and (iii) a combination of both. Finally, to facilitate the modeling of smooth and sharp details found in the estimated force and torque signals, the RCNN is optimized with a loss function designed with the Root Mean Squared Error (RMSE) and Gradient Different Loss (GDL), respectively. The GDL has been investigated in the prediction of future frames from video sequences as discussed in~\cite{Mathieu_2015}, enabling a DNN to render sharp images avoiding blurred pixels. Nonetheless, this concept has neither been extended nor studied for the prediction of time-varying signals. 
%
%

Although models based on CNN and LSTM networks have been investigated in different domains (as discussed in Section~\ref{section:intro:dnn_for_processing_video_sequences}), their application to the force estimation task comes with its own challenges. Therefore, two important goals of this research work are: (i) to reveal the advantages and downsides of a force estimation model based on deep neural networks, and (ii) define future research directions for its implementation on real scenarios. To this end, the following contributions are made:
\begin{itemize}
	\setlength\itemsep{0em}
	\item A RCNN model is proposed for the estimation of interaction forces between tool and tissue relying on a single camera. This method has potential applications in scenarios where a stereo vision system is unavailable, and consequently, depth information.
	\item The effectiveness of applying transfer learning techniques is investigated with the objective of finding a compact feature vector representation for every video frame. For this purpose, the pre-trained VGG16 network~\cite{Simonyan_2015} in the ImageNet dataset is used. This approach allows encoding complex phenomena described in video sequences, such as the deformation of tissues and specular reflections, in a feature vector representation automatically learned from data. This representation is easier to process by a model that learns sequences of data, such as an LSTM network.
	\item A loss function designed with the RMSE and GDL is investigated to facilitate the modeling of smooth and sharp details found in force/torque signals. This loss function composition provides more accurate force estimations than considering only RMSE during the RCNN optimization.
	\item Video pre-processing techniques, specifically mean frame removal and space-time transformations, discussed in~\cite{Pfister_2014} and~\cite{Owens_2015} respectively, were studied to ease the learning process of the RCNN. Mean frame removal was found useful to discard those regions in video sequences which do not contribute to the learning process, such as the static background. The space-time transformation, allows emphasizing motion produced by tool-tissue interactions, in a new image representation created from three consecutive frames.
\end{itemize}
The next sections are organized as follows. Section~\ref{section:problem_statement} defines the problem statement. Section~\ref{section:ramis_system_dataset} describes the dataset acquisition using an experimental robotic platform, and the pre-processing operations applied to this data. Section~\ref{section:force_estimation_model} details the proposed RCNN architecture for force estimation. Section~\ref{section:experiments} presents the experiments, providing details related to the two stage RCNN optimization, and describes how the robustness of the RCNN model was evaluated. Section~\ref{section:results} discusses the results of the experiments and analyses the quality of the estimated force signals with different metrics. Finally, Section~\ref{section:conclusion} presents the conclusions and future work.

%
%
\section{Problem Statement}
\label{section:problem_statement}
Given sequences of video frames $X_t^{video}\in\Re^{h\times w\times c}$ ($h$, $w$ and $c$ stand for image height, width and number of channels, respectively) and tool data $X_t^{tool}\in\Re^8$, the objective is to find a non-linear model $\mathcal{F}(.)$ with parameters $\mathcal{W}$, that maps $X_t^{video}$ and $X_t^{tool}$
to sequence of estimated forces $\widehat{Y}_t\in\Re^6$ at each time instant $t$, as expressed in Equation (\ref{eq:non_linear_mapping}). 
The elements of the input vector $X_t^{tool}$ are shown in Equation (\ref{eq:tool_signals_vector}), where $P^{tool}_t=[x_t, y_t, z_t]$ is a vector describing the tool-tip trajectory in the 3D space, $\Lambda^{tool}_t = [u_t, v_t, w_t]$ is an unitary vector that defines the tool orientation in 3D space (coincident with the tool-axis direction), $\theta_t$ is the angle of rotation around this axis, and $s_t$ is the tool grasper status, defined in Equation (\ref{eq:tool_status}). The tool-tip trajectory ($P^{tool}_t$) and its orientation (defined by $\Lambda^{tool}_t$ and $\theta_t$) are illustrated in Fig.~\ref{fig:experimental_setup}. 
The elements of the output vector $\widehat{Y}_t$ (shown on the left of Equation~(\ref{eq:non_linear_mapping})) are the estimated forces, $\hat{F}_t=[\hat{f}_t^x, \hat{f}_t^y, \hat{f}_t^z]$, and torques, $\hat{T}_t=[\hat{\tau}_t^x, \hat{\tau}_t^y, \hat{\tau}_t^z]$, in the 3D space. Thus, $\widehat{Y}_t=[\hat{F}_t,\hat{T}_t]'=[\hat{f}_t^x, \hat{f}_t^y, \hat{f}_t^z, \hat{\tau}_t^x, \hat{\tau}_t^y, \hat{\tau}_t^z]'$.

In the present work, $\mathcal{F}( . )$ is learnt from data by using a deep neural network. Therefore, given a rich dataset $\mathcal{D}$ consisting of video sequences $X_t^{video}$, tool data $X_t^{tool}$ and ground-truth interaction forces $Y_t$, the goal is to find the parameters $\mathcal{W}$ that satisfy Equation (\ref{eq:non_linear_mapping}) in the context of an optimization framework. A causal constraint is enforced, that is, a estimated force vector $\widehat{Y}_t$ at the current time step, is computed by processing samples from $X_t^{video}$ and $X_t^{tool}$ at the current and previous time steps (i.e.~$t$, $t-1$, $t-2$, $t-3$, ...). In the reported methodology and experiments, the tool orientation remained fixed, therefore, $X_t^{tool}=[P^{tool}_t, s_t]=[ x_t, y_t, z_t, s_t ]'\in \Re^4$. Nonetheless, in the general case, the full vector $X_t^{tool} \in \Re^8$ should be considered.
\begin{equation}\label{eq:non_linear_mapping}
\text{\footnotesize
	$\widehat{Y}_t = \mathcal{F} ( X_t^{tool}, X_t^{video}; \mathcal{W} )$
}
\end{equation}
\begin{equation}\label{eq:tool_signals_vector}
\text{\footnotesize
	$X_t^{tool}=[ P^{tool}_t, \Lambda^{tool}_t, \theta_t, s_t ]'$
}
\end{equation}
\begin{equation} 
\label{eq:tool_status}
\text{\footnotesize
	$s_t = \begin{cases} 
	1 & \text{If} \text{ the grasper is open.}\\
	0 & \text{If} \text{ the grasper is closed.}\\
	\end{cases}$
}
\end{equation}

%
%
\section{Dataset Acquisition \& Pre-processing}
\label{section:ramis_system_dataset}
\begin{figure}[!t]
	\centering
	\includegraphics[width=\columnwidth]{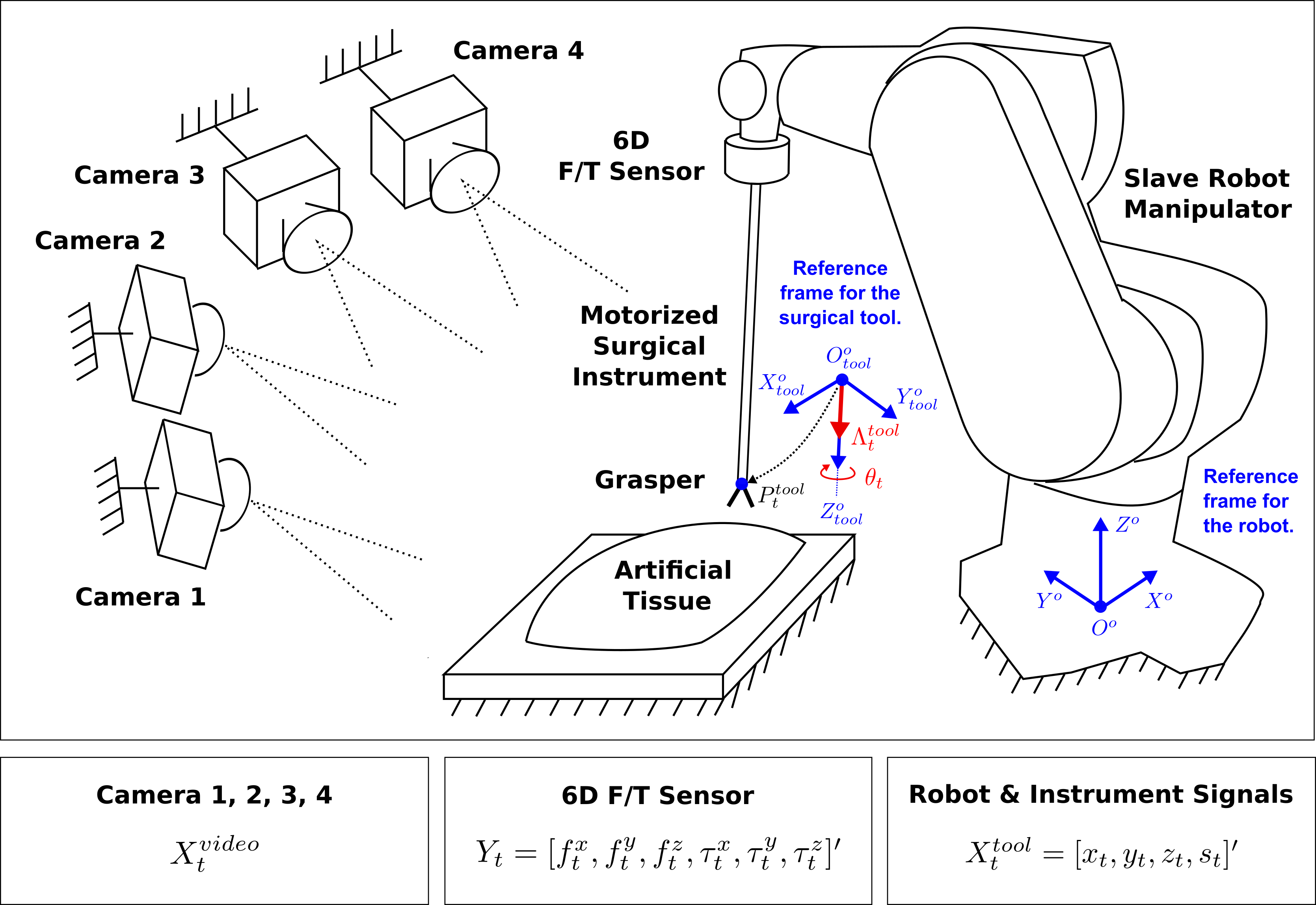}
	\caption{
		Diagram of the experimental setup used to create the dataset. In the bottom, the three blocks relate devices/sensors to the recorded data (in vector form). The reference frame assigned to the robot is $O^{o}=\{X^o, Y^o, Z^o\}$, while the reference frame of the surgical tool-tip with respect to the robot is $O^{o}_{tool}=\{X^{o}_{tool}, Y^{o}_{tool}, Z^{o}_{tool}\}$. 
		The origin of the reference frame $O^{o}_{tool}$, is located at the tool-tip. Therefore, it describes the tool-tip trajectory at each time instant $t$, $P^{tool}_t=[x_t, y_t, z_t]$. The tool orientation, defined by the vector $\Lambda^{tool}_t=[u_t, v_t, w_t]$ and scalar $\theta_t$, was fixed during the experiments. The vector $\Lambda^{tool}_t$ is aligned with the $Z^{o}_{tool}$ axis, which is colinear with the tool shaft.
	}
	\label{fig:experimental_setup}
\end{figure}
\begin{figure}[!t]
	\centering
	\includegraphics[width=\columnwidth]{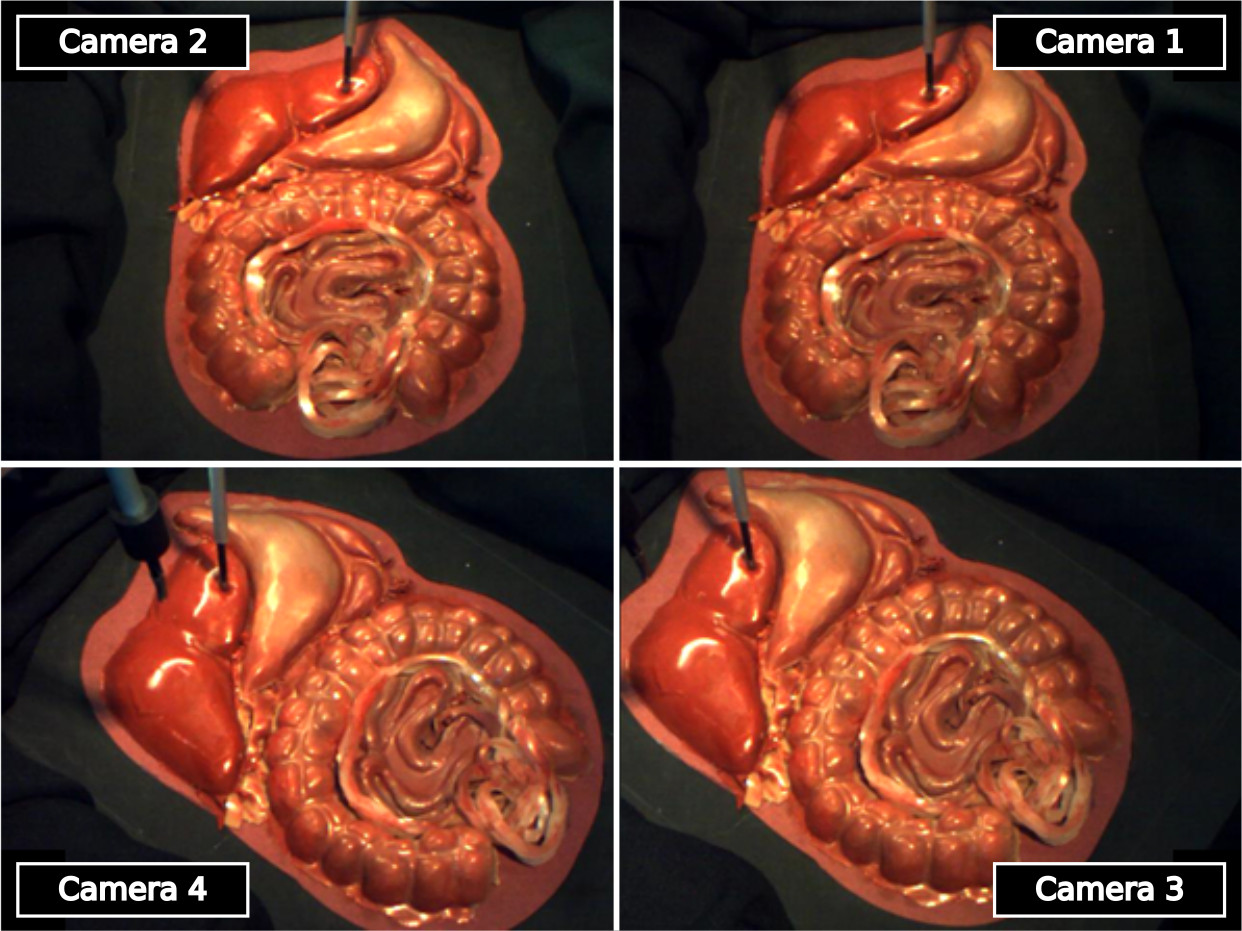}
	\caption{A sample of video frames recorded by the four synchronized cameras. The tool is performing a pushing task over the artificial organs (digestive apparatus).}
	\label{fig:sample_of_video_frames}
\end{figure}

Due to the lack of public datasets related to the application of VBFS in RAMIS, an experimental platform was designed to evaluate the proposed approach, as depicted in Fig. \ref{fig:experimental_setup}. This platform was used to record video sequences, tool data, and ground-truth interaction forces:
\begin{itemize}
	\setlength\itemsep{0em}
	\item \textbf{Video Sequences.} A collection of 44 video sequences, totaling 4.31 hours, were recorded using 4 digital cameras (DFK 72BUC02) with the objective to provide rich visual information from different perspectives. The four cameras were synchronized and the video sequences were recorded with a resolution of $480 \times 640$ pixels at 50 frames per second, in RGB color space. The target scenario consists in a motorized surgical instrument with grasping capability, mounted on a slave robot manipulator (St\"{a}ubli RX60B) that interacts with a digestive apparatus made of artificial tissue (Silicone-Smooth On ECOFLEX 0030). A sample of frames captured by the 4 cameras illustrates the aforesaid scenario in Fig. \ref{fig:sample_of_video_frames}. They show specularities and highlights rendered on the artificial tissue surface, a phenomenon that is present in real minimally invasive surgery scenarios.
	\item \textbf{Tool Data.} The tool-tip trajectory in the 3D space ($P^{tool}_t=[x_t, y_t, z_t]$) and the tool grasping status ($s_t$) were provided, at each time instant, by the slave robot manipulator and the motorized surgical instrument, respectively.
	\item \textbf{Ground-Truth Force.} The interaction forces and torques between the surgical instrument tip and artificial tissue were acquired by a 6D force/torque sensor (ATI Gamma SI-32-2.5) with its $z$ axis aligned with the surgical instrument shaft. The measured forces lie in the range +2.5/-10 N and the torques in +/- 5 Nm, which are consistent with those values reported in a real scenario \cite{Picod_2005}. 
\end{itemize}

Thereafter, a series of pre-processing operations were applied to the tool data, ground-truth interaction force and video frames. The pre-processing of the tool-tip trajectory $P^{tool}_t=[x_t, y_t, z_t]$, was carried out by removing the mean and subsequently scaling its amplitude to the range +/- 1. The grasper status $s_t$ does not need any processing. The ground-truth interaction forces $Y_t$, were compensated with an offset and scaled to the range +/- 5. Additional processing steps, such as time shifting and re-sampling, were applied to the aforesaid signals to synchronize them with the video frames. Moreover, filtering techniques were applied to remove noise.

\begin{figure}[!t]
	\centering
	\includegraphics[width=\columnwidth]{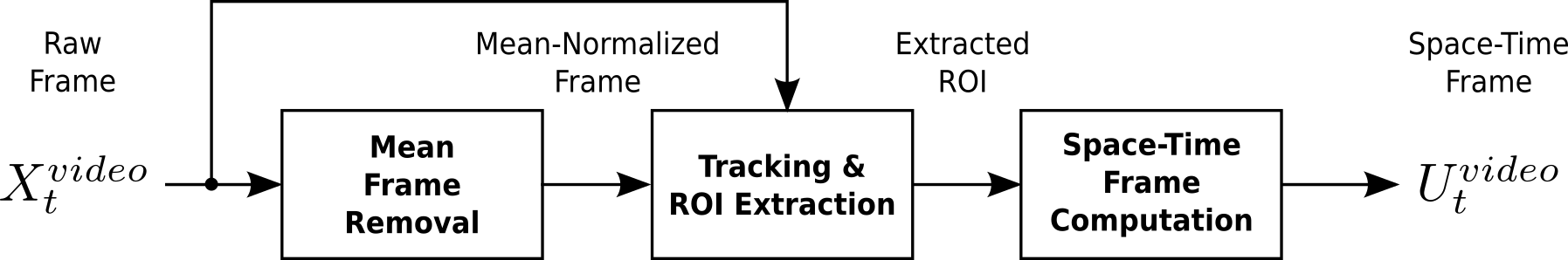}
	\caption{Block diagram of the pre-processing steps applied to video frames.}
	\label{fig:video_preprocessing_block_diagram}
\end{figure}

Video frames required more elaborated pre-processing steps, which can be summarized in the block diagram shown in Fig. \ref{fig:video_preprocessing_block_diagram}, where $X_t^{video}$ and $U_t^{video}$ represent the raw and pre-processed video frames, respectively. Each operation in the block diagram was implemented using OpenCV~\cite{opencv_library} and is described as follows:
\begin{enumerate}
	\setlength\itemsep{0em}
	\item \textbf{Mean Frame Removal}. A mean frame was computed for every video sequence by averaging all the raw frames (with equal contribution). Subsequently, a subtraction operation was performed over the RGB channels, by removing the corresponding mean frame from all the raw frames in the corresponding video sequence. During this process, the pixel values were scaled properly, to conserve negative values. In~\cite{Pfister_2014}, this method was shown to reduce over-fitting of CNNs due to static background present in video sequences. 
	\item \textbf{Tracking of Regions of Interest}. To provide meaningful visual information to the proposed network, a region of interest of dimensions $200\times300$ pixels, corresponding to the area of interaction between tool and tissue, was tracked and extracted from every mean-normalized frame ($480\times640$ pixels). This operation was carried out by processing mean-normalized and raw frames. The result is a mask of foreground pixels describing image regions where tool-tip motion is present. For this purpose, standard computer vision techniques were used, including image filtering (for image noise reduction), foreground extraction (to compute the mask of foreground pixels) and morphological operations (to refine the mask of foreground pixels). 
	\item \textbf{Space-Time Frame Transformation}. This transformation, described in~\cite{Owens_2015}, is applied over the extracted regions of interest with the objective to model tool motion and tissue deformation. It represents an alternative method to the optical flow, which is computationally more expensive. A space-time frame is defined by the previous, current and next RGB frames, each one converted to grayscale. During the experiments, this operation was carried out by concatenating these three frames only every 15 samples. This undersampling is due to the high frame rate of the cameras and the slow motion of the surgical tool.
\end{enumerate}

\begin{figure}[!t]
	\centering
	\begin{tabular}{c}
		\subfloat[Pushing Task]{\includegraphics[width=\columnwidth]{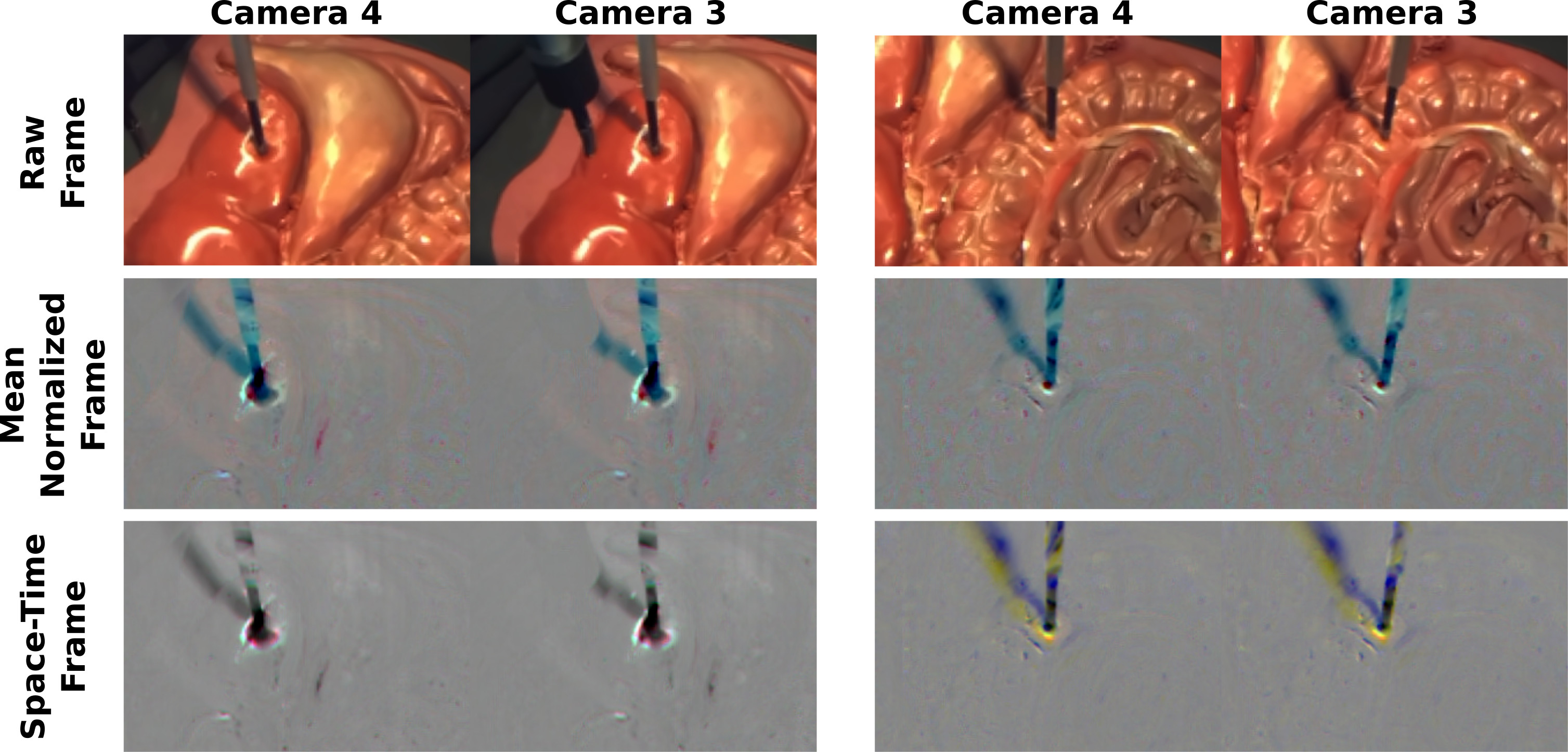}
			\label{fig:sample_of_preprocessed_video_frames:pushing}}\\
		\subfloat[Pulling Task]{\includegraphics[width=\columnwidth]{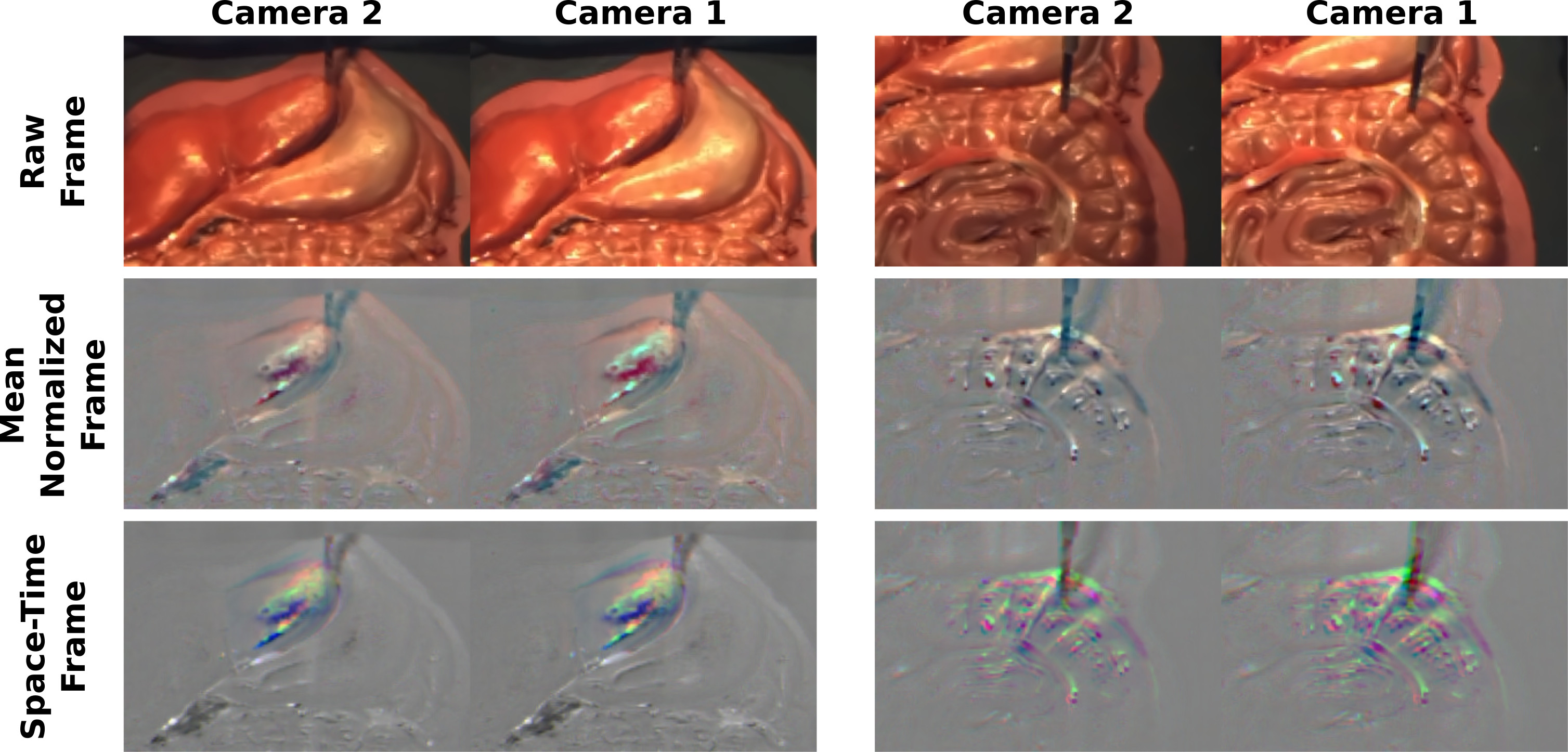}
			\label{fig:sample_of_preprocessed_video_frames:pulling}} 
	\end{tabular}
	\caption{A sample of raw video frames after the mean frame has been removed and the space-time transformation has been applied, for each surgical task.}
	\label{fig:sample_of_preprocessed_video_frames}
\end{figure}

A comparison between regions of interest extracted from the raw, mean-normalized and space-time frames is presented in Fig. \ref{fig:sample_of_preprocessed_video_frames}, for each surgical task. The last row of Fig.~\ref{fig:sample_of_preprocessed_video_frames:pushing} and Fig.~\ref{fig:sample_of_preprocessed_video_frames:pulling} shows that both, tool motion and tissue deformation are emphasized in the space-time domain, and specular reflections are partially suppressed.

%
%
\section{Force Estimation Model}
\label{section:force_estimation_model}
\begin{figure*}[!th]
	\centering
	\includegraphics[width=0.925\textwidth]{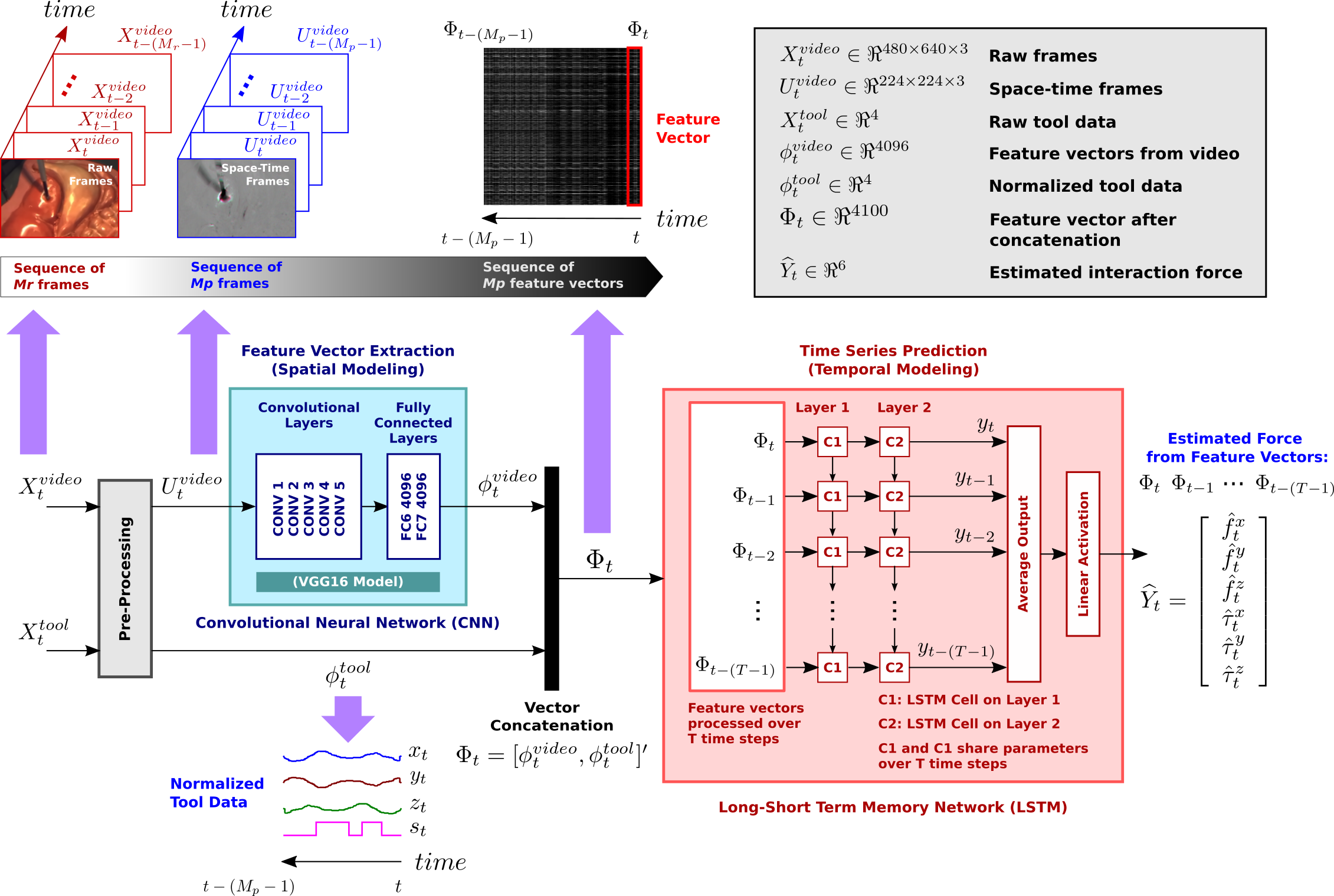}
	\caption{The RCNN architecture consists in a CNN serially connected with an LSTM network. First, pre-processing operations are applied to the input data consisting of raw video sequences ($X_t^{video}$) and tool data ($X_t^{tool}$). Therefore, a sequence of raw data ($X_t^{video}$ and $X_t^{tool}$) of size $M_r$ is transformed into a new sequence of pre-processed data ($U_t^{video}$ and $\phi^{tool}_t$, respectively) of size $M_p$, where $M_p < M_r$. The size difference of these two sequences results from the space-time transformation applied to raw video frames, which is computed by concatenating three consecutive (grayscale) frames spaced in time (in the experiments this spacing correspond to 15 frames). Subsequently, the CNN extracts feature vectors ($\phi_t^{video}$) from the pre-processed input video sequence ($U_t^{video}$). Afterwards, these feature vectors ($\phi_t^{video}$) and the normalized tool data ($\phi_t^{tool}$) are concatenated, resulting in a new feature vector ($\Phi_t$). Finally, these new feature vectors ($\Phi_t$) are fed into the LSTM network, which models their temporal structure to render the estimated force as output ($\widehat{Y}_t$).}
	\label{fig:proposed_rcnn_architecture}
\end{figure*}

The processing of video sequences requires taking into account their spatial and temporal information. In contrast, for tool data only the temporal information is relevant. To deal with these two kinds of data, a Recurrent Convolutional Neural Network (RCNN) architecture is proposed to carry out the force estimation task. This RCNN consists of a Convolutional Neural Network (CNN) serially connected with an Long-Short Term Memory (LSTM) network. Each neural network has a specific objective and is optimized separately as described below:
\begin{enumerate}
	\setlength\itemsep{0em}
	\item \textbf{CNN Optimization.} The modeling of the spatial component present in video sequences is carried out by the CNN. This neural network is optimized for a regression task on the dataset. The input and output data consist of space-time frames (in RGB color space with a resolution of $224\times224$ pixels) and interaction forces (6-dimensional force vectors), respectively. Subsequently, the CNN is used as a feature extractor. Its objective is to find a compact feature vector representation  (a 4096-dimensional vector) that encodes high-level abstractions of the input data (space-time frames).
	\item \textbf{LSTM Network Optimization.} The temporal information present in tool data and feature vectors computed by the CNN is modeled by the LSTM network. This neural network is optimized for a regression task, by taking the tool data (4-dimensional vectors) and feature vectors (4096-dimensional vectors) as input with the objective of estimating a tool-tissue interaction force vector (a 6-dimensional vector) at each time instant. 
\end{enumerate}

The RCNN architecture is depicted in Fig.~\ref{fig:proposed_rcnn_architecture}. This illustration shows the flow of data from the input to the output in four stages. First, pre-processing operations are applied to the raw video sequences $X_t^{video}\in\Re^{200\times300\times3}$ and tool data $X_t^{tool}\in\Re^4$, resulting in $U_t^{video}\in\Re^{224\times224\times3}$ and $\phi_t^{tool}\in\Re^{4}$, respectively. Second, the CNN extracts feature vectors, $\phi_t^{video}\in\Re^{4096}$, from the pre-processed input video sequence, $U_t^{video}$. Third, these feature vectors ($\phi_t^{video}$) and the normalized tool data ($\phi_t^{tool}$) are concatenated, resulting in a new feature vector, $\Phi_t\in\Re^{4100}$. Finally, these new feature vectors are fed into the LSTM network, which models their temporal structure to render the estimated force $\widehat{Y}_t\in\Re^6$ as output. 

\begin{figure*}[!thpb]
	\centering
	\includegraphics[width=0.95\textwidth]{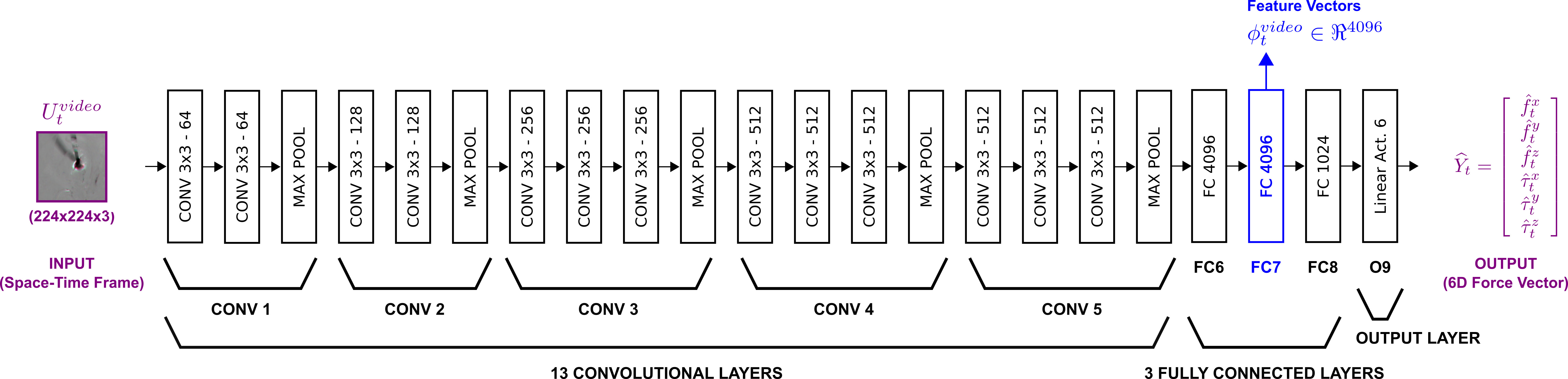}
	\caption{VGG16 network~\cite{Simonyan_2015} used for fine-tuning and feature vector extraction. It consists of 13 convolutional (kernel size of $3\times3$) and 3 fully-connected layers. In this illustration, the convolutional layers are grouped into CONV 1, ..., CONV 5. The fully connected layers are referred to as FC6, FC7, and FC8. The Rectifier Linear Unit (ReLU) is used as activation function in all layers except the output layer, O9, which is densely connected with a linear activation. The number of output feature maps for each convolutional layer and the size of each fully connected layer are indicated with the last number inside the corresponding layer. At test time, the feature vectors $\phi_t^{video}\in\Re^{4096}$, are extracted from the layer FC7 (shown in blue color).
	}
	\label{fig:vgg16_network_architecture}
\end{figure*}

For the task of feature vector extraction from video sequences, the pre-trained VGG16 network model proposed in~\cite{Simonyan_2015}, in the context of image classification, was fine-tuned on the dataset. Specifically, in this process, the VGG16 network computes a force vector as output conditioned on an input video frame, while the network's parameters, in all layers, are adjusted in the context of an optimization framework. During the fine-tuning process, generic features (i.e.~computed in the first and second layers) are less prone to change, while specific features (i.e.~computed towards the last layer) will be adjusted according to the force estimation dataset. The VGG16 network, shown in Fig.~\ref{fig:proposed_rcnn_architecture} as the block in blue color, is detailed in Fig.~\ref{fig:vgg16_network_architecture}. To match the neural network output size with that of the force vectors, the softmax layer of dimension 1000 (found in the original VGG16 network), was replaced by a densely connected layer of dimension 6 (with linear activation). The space-time frames ($U_t^{video}$) were resized preserving their aspect ratio (by centered cropping and resampling operations), from $200\times300$ to $224\times224$ pixels (matching the network's input size). After the fine-tuning process is completed, the feature vectors $\phi_t^{video}$, are extracted from the fully-connected layer FC7 (shown in Fig.~\ref{fig:vgg16_network_architecture} in blue color).

\subsection{Loss Function Design}
\label{subsection:force_estimation_model:cnn_optimizaton}
The loss function has an important impact in the design of deep neural networks applied to regression tasks. This impact is also extended to the design of regression models based on CNNs. For instance, human pose estimation was studied in \cite{Pfister_2014} with a CNN optimized with the standard $L2$ loss function (sensitive to outliers) to penalize the distance between predicted and ground-truth upper-body joint positions. The same application was investigated in \cite{Belagiannis_2015}, by minimizing Tukey's bi-weight function to achieve robustness against outliers. Recently, \cite{Mathieu_2015} proposed a method for predicting future images from a video sequence by the minimization of a loss function that takes into account the Gradient Different Loss (GDL). This method allows overcoming the prediction of blurry images when only the mean squared error is considered in the loss function. In the present work, the GDL has been extended to the estimation of time-varying force signals. Therefore, each network (CNN and LSTM), that defines the proposed RCNN architecture was optimized separately with a loss function composed of the Root Mean Squared Error (RMSE), and the GDL. The RMSE penalizes the distance between estimated and ground-truth 6D force vectors, while the GDL the distance between their gradients. Intuitively, the RMSE and GDL ease the modeling of smooth and sharp details found in force/torque signals, respectively.

The loss function discussed above, denoted as $\mathcal{L}\in\Re$, is mathematically expressed in Equation (\ref{eq:loss}), where $\alpha \in [0, 1]$ represents a trade-off between the RMSE ($\mathcal{L}_{RMSE}\in\Re$) and GDL ($\mathcal{L}_{GDL}\in\Re$). The RMSE expressed in Equation (\ref{eq:loss_rmse}) computes the distance between the ground-truth $Y_i^{(j)}\in\Re$ and the estimated $\widehat{Y}_i^{(j)}\in\Re$ force components, where $i$ indexes the samples in the dataset $\mathcal{D}$ and $j$ the $N$ force components. In this equation, $\rho(x_i)\in\Re$ is a function applied to the scalar $x_i\in\Re$, which is computed for the $i$-th sample in the dataset. The parameters described for the RMSE are also found in the GDL expressed in Equation~(\ref{eq:loss_gdl}).

As mentioned in the beginning of this section, the RCNN optimization consists in two stages. In the first stage, the VGG16 network (shown in Fig.~\ref{fig:vgg16_network_architecture}) is fine-tuned with a loss function defined in Equations (\ref{eq:loss})-(\ref{eq:loss_gdl}). This neural network $\mathcal{F}_{1}$ with parameters $\mathcal{W}_{1}$, is represented by Equation (\ref{eq:vgg16_prediction}), where $\widehat{Y}_i\in\Re^{N}$ stands for the estimated force vector, given as input the $i$-th space-time frame, $U_i^{video}$. In the subsequent stage, the LSTM network $\mathcal{F}_{2}$ with parameters $\mathcal{W}_{2}$ shared across $T$ time steps, is trained using the same loss function. This neural network is expressed in Equation (\ref{eq:lstm_prediction}). It outputs $\widehat{Y}_i\in\Re^{N}$, that is, the estimated force vector at the time instant $i$, given as input a sequence of $T$ feature vectors $\Phi_d$, at time steps $d = i, i-1, i-2,..., i-(T-1)$, (see the LSTM network depicted in Fig.~\ref{fig:proposed_rcnn_architecture}).

The selection of $\rho(x_i)$ in Equation (\ref{eq:loss_rmse}) and (\ref{eq:loss_gdl}), was different for each optimization step. Motivated by the work in \cite{Owens_2015}, the VGG16 network was fine-tuned with the logarithmic function stated in Equation (\ref{eq:vgg16_function_rho}), where the index $i$ is omitted for clarity in the notation, $\gamma\in\Re$ is a parameter, and $\epsilon$ a small positive constant (which avoids the evaluation of the logarithmic function at zero). This function saturates large gradients produced by the error between ground-truth and estimated data, adding robustness to the optimization. Equation (\ref{eq:vgg16_function_rho}) was applied to (\ref{eq:loss_rmse}) using $\gamma=2.0$, resulting in a function that operates over the mean squared differences between ground-truth and estimated data. In contrast, Equation (\ref{eq:vgg16_function_rho}) was applied to (\ref{eq:loss_gdl}) with $\gamma=1.0$, resulting in a function that process the absolute difference of residuals. Another design choice for $\rho(x_i)$ consist of a linear function, shown in Equation (\ref{eq:lstm_function_rho}) (where the index $i$ is omitted), which provides better convergence during the LSTM network optimization.
\begin{equation}\label{eq:loss}
\text{\footnotesize
	$\mathcal{L} = \alpha\thinspace\mathcal{L}_{RMSE} + (1-\alpha)\thinspace\mathcal{L}_{GDL}$
}\end{equation}
\begin{equation}\label{eq:loss_rmse}
\text{\footnotesize
	$\mathcal{L}_{RMSE} = \sum_{i=1}^{|\mathcal{D}|} \rho (x_i),~x_i=\sqrt{ \frac{1}{N} \sum_{j=1}^{N} (Y_i^{(j)} - \widehat{Y}_i^{(j)})^{2} }$
}\end{equation}
\begin{equation}\label{eq:loss_gdl}
\text{\footnotesize
	$\mathcal{L}_{GDL} = \sum_{i=1}^{|\mathcal{D}|} \rho (x_i),~x_i=\sum_{j=1}^{N} \Big| |Y_i^{(j)} - Y_{i-1}^{(j)}| - |\widehat{Y}_i^{(j)} - \widehat{Y}_{i-1}^{(j)}| \Big|$
}\end{equation}
\begin{equation}\label{eq:vgg16_prediction}
\text{\footnotesize
	$\widehat{Y}_i = \mathcal{F}_{1}(U_i^{video}; \mathcal{W}_{1})$
}\end{equation}
\begin{equation}\label{eq:lstm_prediction}
\text{\footnotesize
	$\widehat{Y}_i = \mathcal{F}_{2}(\Phi_d; \mathcal{W}_{2})$
}\end{equation}
\begin{equation} 
\label{eq:vgg16_function_rho}
\text{\footnotesize $\rho(x) = \ln{ (x^{\gamma} + \epsilon) }$}
\end{equation}
\begin{equation}
\label{eq:lstm_function_rho} \text{\footnotesize $\rho(x) = x$}
\end{equation}

%
%
\section{Experiments}
\label{section:experiments}
The proposed RCNN architecture was implemented in Python using the Tensorflow~\cite{Abadi_2016} framework. The experiments were carried out using multiple graphics processing units, including the NVIDIA Titan X and Tesla K80. The dataset samples were split into the training and test sets, as detailed in Table~\ref{table:dataset_composition}.
\subsection{Experiments Design}
First, the VGG16 network is fine-tuned with the objective to find a feature vector representation $\phi_t^{video} \in \Re^{4096}$, for every space-time frame $U_t^{video} \in \Re^{224\times224\times3}$ (see Fig.~\ref{fig:vgg16_network_architecture}). Subsequently, in the LSTM network optimization, three types of feature vectors $\Phi_t$ (processed at every time step $t$), were evaluated as input data:
\begin{itemize}
	\setlength\itemsep{0em}
	\item \textbf{Case I}. Only tool data as input: $\Phi_t=\phi_t^{tool}\in\Re^4$.
	\item \textbf{Case II}. Only feature vectors extracted from video sequences as input: $\Phi_t=\phi_t^{video}\in\Re^{4096}$.
	\item \textbf{Case III}. Both, tool data and feature vectors extracted from video sequences as input: $\Phi_t=[\phi_t^{video},\phi_t^{tool}]'\in\Re^{4100}$.
\end{itemize}
For each aforesaid case, two loss functions were evaluated to investigate the contribution of the RMSE and GDL terms that appear in Equation (\ref{eq:loss}):
\begin{itemize}
	\setlength\itemsep{0em}
	\item \textbf{Loss A}. Setting $\alpha=0.75$ results in the loss $\mathcal{L}=0.75\thinspace\mathcal{L}_{RMSE}+0.25\thinspace\mathcal{L}_{GDL}$. Thus, more importance is given to the RMSE than to the GDL, due to the faster convergence of the former term compared to the latter.
	\item \textbf{Loss B}. Setting $\alpha=1.0$ results in the loss $\mathcal{L}=\mathcal{L}_{RMSE}$. Therefore, only the RMSE is considered in the optimization. 
\end{itemize}
Therefore, a total of six cases, following the format~\emph{case number-loss type}, were analyzed during the LSTM network optimization. These cases are referred to as I-A, I-B, II-A, II-B, III-A, and III-B. The optimization of the VGG16 and LSTM networks is detailed in Sections~\ref{subsection:experiments:vgg16_optimization} and~\ref{subsection:experiments:lstm_optimization}, respectively. In Section~\ref{subsection:experiments:rcnn_model_robustness}, additional experiments are described, whose objective is to evaluate the robustness of the proposed RCNN model. Finally, Section~\ref{subsection:experiments:rcnn_vs_armax} explains an experiment in which a time-series model is studied in the force estimation task.

\begin{table}[!t] 
	\renewcommand{\arraystretch}{1.30} 
	\caption{Dataset samples used in the experiments: (a) Complete dataset including both, pushing and pulling tasks, (b) dataset describing only pushing and (c) pulling tasks.} 
	\label{table:dataset_composition}
	\centering
	\scriptsize
	\begin{threeparttable}[t]
\begin{tabular}{|l|cc|c|c|}
	\hline
	\bfseries Dataset		& \multicolumn{2}{c|}{\bfseries Video Sequences}	        & \bfseries Samples\tnote{(1)}~    & \bfseries Percentage \\ 
	\cline{2-3}
	\bfseries Type      	& \bfseries \# Files			& \bfseries Duration\tnote{(2)}~  &  										&                      \\
	\hline
	\multicolumn{5}{|c|}{(a) Complete Dataset}  \\
	\hline
	Training	 			& 28 							& $\sim$3 h 19 min			& 597388		& 77~\% \\
	Test					& 16 							& $\sim$1 h					& 179292		& 23~\%  \\
	\hline
	\bfseries Total			& \bfseries 44 					& \bfseries $\sim$4 h 19 min	& \bfseries 776680	& \bfseries 100~\% \\
	\hline
	\multicolumn{5}{|c|}{(b) Pushing Tasks}  \\
    \hline
	Training				& 16 							& 106.26 min 				& 318776			& 41~\% \\		
	Test					& 12 							& 46.48	min					& 139448			& 18~\% \\
	\hline 	
	\bfseries Total			& \bfseries 28					& \bfseries 152.74 min  	& \bfseries 458224 	& \bfseries 59~\% \\
	\hline
	\multicolumn{5}{|c|}{(c) Pulling Tasks}  \\
	\hline
	Training	 			& 12 							& 92.87 min					& 278612			& 36~\% \\
	Test					& 4 							& 13.28 min					& 39844				& 5~\%  \\
	\hline
	\bfseries Total			& \bfseries 16 					& \bfseries 106.15 min		& \bfseries 318456	& \bfseries 41~\% \\
	\hline
\end{tabular}
\begin{tablenotes}
	\item[(1)] Each sample consists of a video frame (224$\times$224$\times$3), a 4D tool data vector, and a 6D ground-truth force vector.
	\item[(2)] Computed as $T = N / F_r$, where $T$ is the video duration, $N$ the total number of frames, and $F_r$ is the frame rate (50 frames per second).
\end{tablenotes}
\end{threeparttable}%

\end{table}

\subsection{VGG16 Network Fine-tuning}
\label{subsection:experiments:vgg16_optimization}
\begin{figure}[!t]
	\centering
	\includegraphics[width=1.0\columnwidth]{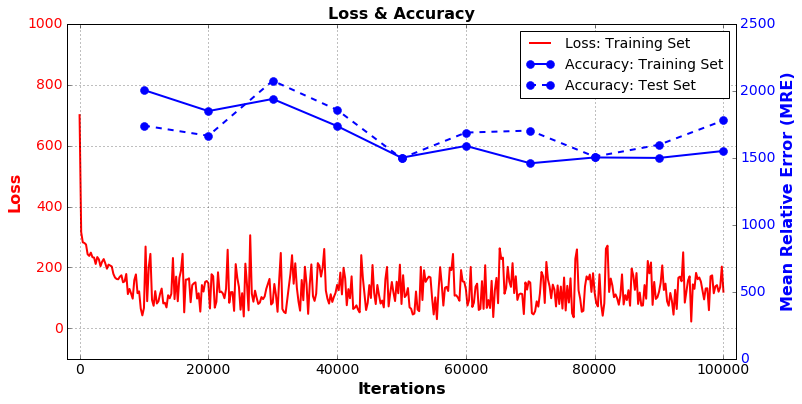}
	\caption{Computed loss (in red) and accuracy (in blue), during the fine-tuning of the VGG16 network. }
	\label{fig:vgg16_optimization:loss_and_accuracy}
\end{figure}

\begin{figure}[!t]
	\centering
	\includegraphics[width=1.0\columnwidth]{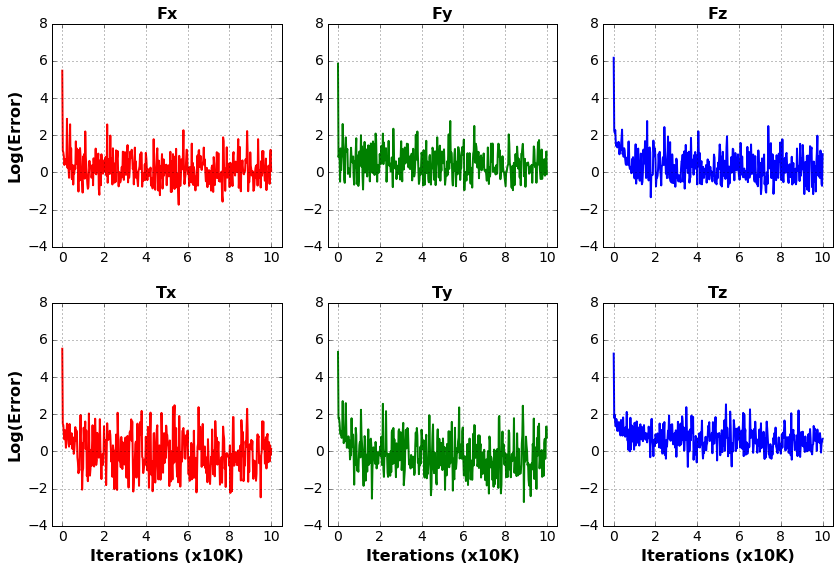}
	\caption{Logarithm of the error per force component computed (on data in the training set) during the fine-tuning process.}
	\label{fig:vgg16_optimization:error_per_force_component}
\end{figure}

The VGG16 model, with weights pre-trained on the ImageNet dataset~\cite{Russakovsky_2015}, was fine-tuned over 100K iterations with the RMSProp optimizer~\cite{RMSProp:Tieleman2012}. Its accuracy was calculated based on the Mean Relative Error (MRE) per batch stated by Equation (\ref{eq:vgg16_optimization:mean_relative_error}), where $M$ stands for the number of samples in a batch of data, $N$ represents the number of force components and $\delta$ is the tolerance error. Thus, the MRE was computed for each batch defined in the dataset and averaged over the total number of batches. The $L2$-norm of the error per force component $\lVert r_j \rVert_2$ described by Equation (\ref{eq:vgg16_optimization:error_per_force_component}), where $j = 1, ..., N$, was also taken into account. Table~\ref{table:vgg16_optimization:hyperparameters} summarizes the hyper-parameters used during the VGG16 network fine-tuning, which were adjusted experimentally. In particular, $\alpha$ was set to 0.8 due to the faster convergence of the RMSE compared to the GDL. Finally, $\epsilon$ was set to 1/100 for numerical stability.
\begin{table}[!t] 
	\renewcommand{\arraystretch}{1.30} 
	\caption{Hyperparameters used for the VGG16 model fine-tuning.} 
	\label{table:vgg16_optimization:hyperparameters}
	\centering
	\scriptsize
	\begin{tabular}{|l|c|}
	\hline
	\bfseries Hyperparameter                          							& \bfseries Value   \\
	\hline
	Learning Rate, $\lambda$             							& $1\times10^{-5}$ \\
	Batch Size, $M$                              							& 50 samples       \\
	Dropout (Fully-Connected Layers)             							& 50 \%            \\
	Parameter $\alpha$ in Equation (\ref{eq:loss})                  					& 0.8      \\
	Parameter $\epsilon$ in Equation (\ref{eq:vgg16_function_rho}) 					& 1/100    \\
	Parameter $\delta$ in Equation (\ref{eq:vgg16_optimization:mean_relative_error}) 	& $1\times10^{-3}$ \\
	\hline 
\end{tabular}
\end{table}
To monitor the evolution of the optimization, the loss function defined in Equation (\ref{eq:loss}) and the logarithm of the error per force component expressed by Equation (\ref{eq:vgg16_optimization:error_per_force_component}) were computed every 250 iterations on the training set. The accuracy was evaluated on the training and test sets every 10K iterations. The plot of the loss and accuracy is shown in Fig. \ref{fig:vgg16_optimization:loss_and_accuracy}, while the error per force component computed for data in the training set, is illustrated in Fig. \ref{fig:vgg16_optimization:error_per_force_component}.
\begin{equation}
\label{eq:vgg16_optimization:mean_relative_error}
\text{\footnotesize
	$MRE = \frac{1}{M} \sum_{i=1}^{M} \sum_{j=1}^{N} \frac{ | Y_i^{(j)} - \widehat{Y}_i^{(j)} | }{\delta}, MRE \in \Re$
}
\end{equation}
\begin{equation} 
\label{eq:vgg16_optimization:error_per_force_component}
\text{\footnotesize
	$\lVert r_j \rVert_{2} = \sqrt{ \sum_{i=1}^{M} (Y_i^{(j)} - \widehat{Y}_i^{(j)})^{2} }, \lVert r_j \rVert_{2} \in \Re$
}
\end{equation}
After the VGG16 network was fine-tuned on the video dataset, visual features $\phi_t^{video}$ were extracted from the fully connected layer FC7 (see Fig.\ref{fig:vgg16_network_architecture}), replacing the Rectifier Linear Unit (ReLU) by the Hyperbolic Tangent (Tanh) non-linearity. By applying the Tanh non-linearity, all values present in the feature vectors are squashed between -/+1. This range of values is expected in the feature vectors to be processed by the LSTM network (during both training and inference stages) since the block-input of this network has the Tanh non-linearity as the activation function. Each feature vector computed by the VGG16 network can be interpreted as a learned representation in the low-dimensional space ($\phi_t^{video}\in\Re^{4096}$) for each input video frame that lies in the high-dimensional space ($U_t^{video}\in\Re^{224\times224\times3}$).
\subsection{LSTM Network Optimization}
\label{subsection:experiments:lstm_optimization}
Three models were empirically evaluated in the force estimation task: (i) The vanilla LSTM network (with added peephole connections), (ii) the Coupled-Input Forget Gate (CIFG) variant of the LSTM network (LSTM-CIFG) and (iii) the Gated Recurrent Unit (GRU). In terms of convergence and quality of prediction, the LSTM-CIFG was superior to the vanilla LSTM and GRU networks; the worst results were obtained with the GRU model. Therefore, the LSTM-CIFG network was selected to carry out the experiments and predict interaction forces between surgical instruments and tissues. 

The LSTM-CIFG network was trained with the RMSProp optimizer, using the hyper-parameters listed in Table~\ref{table:lstm_optimization:hyperparameters}. For case I, this neural network was designed with only 64 cell units per layer due to the low dimensionality of the input data ($\phi_t^{tool} \in \Re^4$), avoiding over-fitting in the training set. In contrast, the neural networks designed for cases II and III required higher capacity (i.e.~more parameters) due to the complexity added by the feature vectors ($\phi_t^{video} \in \Re^{4096}$) in the input data. Therefore, these neural networks were designed with 256 cell units per layer. In all the six cases (I-A, ..., III-B), dropout was applied at the output of each layer as a method for regularization to prevent over-fitting (a higher value was set for the case I). For each case and loss function studied, the total number of iterations required to optimize the LSTM-CIFG network is shown in the last row of Table~\ref{table:lstm_optimization:hyperparameters}. The optimization was stopped after observing that the loss value reached a plateau, and there was no visible improvement in test set accuracy.

The quality of the predicted force signals with respect to the ground truth was assessed by considering two metrics, the Root Mean Square Error (RMSE) and the Pearson Correlation Coefficient (PCC).
%
%
\begin{table}[!t] 
	\renewcommand{\arraystretch}{1.30} 
	\caption{Hyperparameters used for the LSTM network optimization.} 
	\label{table:lstm_optimization:hyperparameters}		
	\centering
	\scriptsize
	\begin{threeparttable}[t]
\begin{tabular}{|l|c|c|c|c|c|c|}
	\hline
	\bfseries Case 				   	& \bfseries I & \bfseries II & \bfseries III & \bfseries I & \bfseries II & \bfseries III \\
	\hline
	\bfseries Loss Function 	    & \multicolumn{3}{c|}{\bfseries A\tnote{(1)}} & \multicolumn{3}{c|}{\bfseries B\tnote{(2)}} \\
	\hline
	Number of Layers               	& \multicolumn{6}{c|}{2}         \\
	\cline{2-7}
	Cells per Layer                	& 64 & 256 & 256 & 64 & 256 & 256               \\ 
	\cline{2-7}
	Time Steps                     	& \multicolumn{6}{c|}{64}                 \\
	\cline{2-7} 
	Learning Rate, $\lambda$   	   	& \multicolumn{6}{c|}{0.0025}             \\
	\cline{2-7}
	Batch Size, $M$                	& \multicolumn{6}{c|}{512 samples}         \\
	\cline{2-7}
	Dropout L1\tnote{(3)}         	& 75\% & 25\% & 25\% & 75\% & 25\% & 25\%  \\
	\cline{2-7}
	Dropout L2\tnote{(4)}         	& 75\% & 25\% & 25\% & 75\% & 25\% & 25\%  \\                                
	\cline{2-7}
	Parameter $\delta$~\tnote{(5)} 	& \multicolumn{6}{c|}{$1\times10^{-3}$}  \\
	\cline{2-7}
	Iterations\tnote{(6)}          	& 99.0 & 39.7        & 57.9 & 99.0      & 49.1        & 26.7 \\
	\hline
\end{tabular}
\begin{tablenotes}
	\item[(1)] Loss function A: $\mathcal{L}=0.75\thinspace\mathcal{L}_{RMSE}+0.25\thinspace\mathcal{L}_{GDL}$.
	\item[(2)] Loss function B: $\mathcal{L}=\mathcal{L}_{RMSE}$.
	\item[(3)] Dropout applied to layer 1 (L1).
	\item[(4)] Dropout applied to layer 2 (L2).
	\item[(5)] Equation~(\ref{eq:vgg16_optimization:mean_relative_error})
	\item[(6)] Total number of iterations ($\times$1000).
\end{tablenotes}
\end{threeparttable}%
\end{table}
\subsection{Robustness of the RCNN Model}
\label{subsection:experiments:rcnn_model_robustness}
Two experiments, described below, were carried out to evaluate the robustness of the RCNN model.

In the first experiment, the robustness of the RCNN model against Gaussian noise added on tool data was evaluated. As the noise intensity was strengthened by increasing the variance, the deterioration of the estimated force signal quality was measured with the PCC and RMSE metrics.  

In the second experiment, the RCNN model performance was evaluated by feeding this neural network with input video sequences pre-processed in offline and real-time modes. In offline mode, the whole video sequence is available for computing and applying pre-processing operations on raw frames, namely mean frame removal and space-time transformation. In contrast, in the real-time mode, only the past frames from video sequences can be used to perform such pre-processing operations. In the context of a real-time scenario, the computation of a mean frame followed by its subtraction from a specific video sequence represents a key pre-processing operation that has an impact on the quality of the estimated force signals. Therefore, in the real-time mode, the mean frame was computed by averaging only past frames in a video sequence. On the other hand, in the offline mode, the mean frame was obtained by averaging all the frames in a video sequence (in the experiments described in Sections~\ref{subsection:experiments:vgg16_optimization} and~\ref{subsection:experiments:lstm_optimization}, it was assumed that all video sequences were available offline). Afterward, the quality of the force estimations that resulted from each pre-processing mode was compared. Two samples of video sequences (from the test set) were used in this experiment, each one related to pushing and pulling tasks. This analysis reveals that the RCNN model is suitable for the task of force estimation in real-time. However, there is a small degradation of the quality of the estimated force signals with respect to the offline mode. These results will be discussed in the next section.

\subsection{RCNN Model vs Time Series Model}
\label{subsection:experiments:rcnn_vs_armax}

A simpler method (not based on neural networks) than the proposed RCNN was investigated in the task of force estimation. For this purpose, an Auto-Regressive Moving Average Model with eXogenous Inputs (ARMAX), commonly used in the context of time series modeling and system identification, was selected to model the complex relationship between the input tool data and the output interaction forces. This model was implemented in MATLAB, and its parameters were estimated using a combination of four line search algorithms, specifically, subspace Gauss-Newton, adaptive subspace Gauss-Newton, Levenberg-Marquardt, and steepest descent. After a single optimization step, the algorithm that provides the lowest cost is selected to estimate the model parameters.
%
%
\section{Results \& Discussion}
\label{section:results}

%
%

The results and discussion of the experiments are presented in four sections. First, Section~\ref{subsection:results:estimated_force_signals} describes the results of the LSTM-CIFG network optimization (which outputs the estimated interaction force, $\widehat{Y}_t$, given as input the feature vectors, $\Phi_t$) and discusses the six cases studied (I-A, ..., III-B). Then, Section~\ref{subsection:results:rcnn_model_robustness}, reports the results from the experiments related to the robustness of the RCNN model in the conditions described in Section~\ref{subsection:experiments:rcnn_model_robustness}. Afterwards, Section~\ref{subsection:results:rcnn_vs_armax}, contrasts the force estimation quality of the RCNN model against the ARMAX model. Finally, Section~\ref{subsection:results:challenges_real_scenarios} discusses the key ideas to improve the RCNN model in the context of real applications. All the results shown in Tables~\ref{table:lstm_optimization:metrics_statistics},~\ref{table:rcnn_model_frames_preprocessed_online_vs_offline}, and~\ref{table:rcnn_vs_armax_model} and Figs.~\ref{fig:signal_quality_metrics}-\ref{fig:robustness_against_noise}, were computed using the normalized signals provided by the RCNN, which are dimensionless and in the range +/-5. On the other hand, Table~\ref{table:force_estimation_quality_in_forcetorque_units} shows the force estimation quality, measured with the RMSE, in physical units.

\subsection{Estimated Force Signals}
\label{subsection:results:estimated_force_signals}
\begin{figure*}[!th]
	\centering
	
	\subfloat[Pushing Tasks]{\includegraphics[width=0.925\columnwidth]{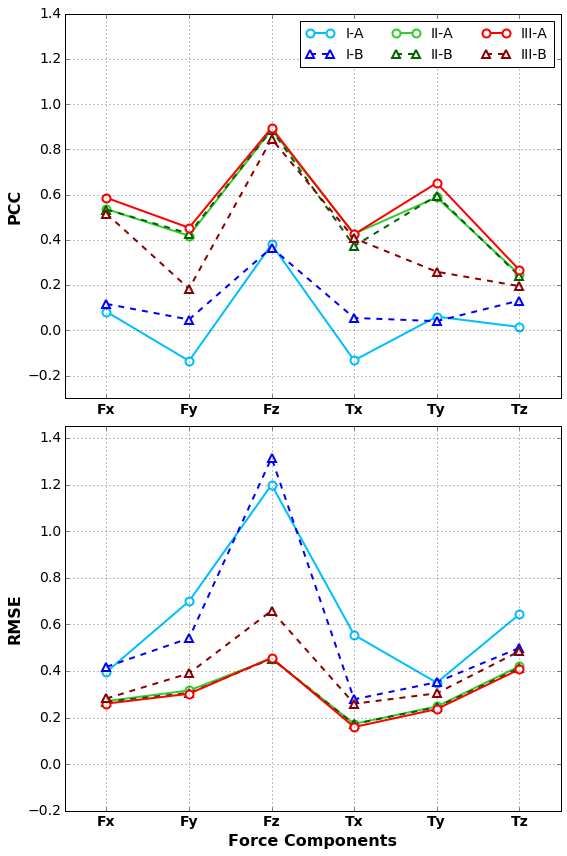}%
		\label{subfig:signal_quality_metrics:pushing}}
	\hfil
	\subfloat[Pulling Tasks]{\includegraphics[width=0.925\columnwidth]{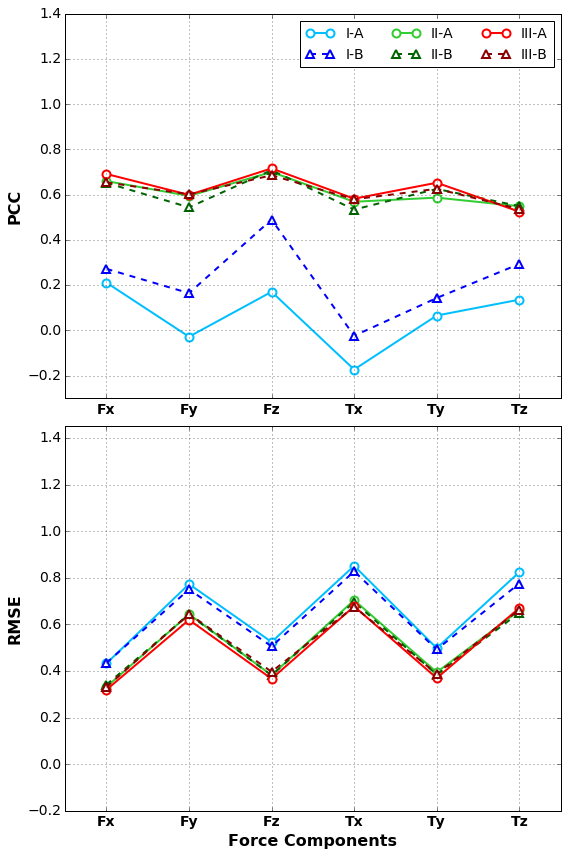}%
		\label{subfig:signal_quality_metrics:pulling}}
	\caption{Force estimation quality measured with the Root Mean Squared Error (RMSE) and Pearson Correlation Coefficient (PCC) for each surgical task, pushing (left column) and pulling (right column) tissue. The six cases studied (I-A, I-B, II-A, II-B, III-A, and III-B) are contrasted in these plots. For the PCC, values closer to 1.0 are better, while for the RMSE values closer 0.0 are desirable. In this illustration, case III-A (solid line in red color) stands out at the best model.}
	\label{fig:signal_quality_metrics}
	
\end{figure*}

After the LSTM-CIFG network optimization was completed, the quality of the estimated force signals (in the test set) was measured with the RMSE and PCC metrics. These metrics are shown in Fig.~\ref{fig:signal_quality_metrics} for each surgical task (pushing and pulling), case (I, II and III) and loss function (loss A and B). From this illustration, case III-A stands out as the best model (solid line in red color), since it has higher PCC values and lower RMSE values with respect to the other cases. On the other hand, the metrics for case III-B (dotted line in dark red color) fall behind those attributed to case III-A in a pushing task (left column), while for pulling tasks (right column) they are close in proximity. For cases II-A (solid line in green color) and II-B (dotted line in dark green color), the PCC and RMSE values are slightly behind the accuracy reported for case III-A. Therefore, the second best model could be either, case II-A or II-B, since their values are very close to each other. Finally, cases I-A (solid line in blue color) and I-B (dotted line in dark blue color), represent the worst models. This conclusion is also justified in Table~\ref{table:lstm_optimization:metrics_statistics}, where the maximum, minimum and mean values of the metrics displayed in Fig.~\ref{fig:signal_quality_metrics} are presented (the best values are highlighted in bold).

The results presented in Fig.~\ref{fig:signal_quality_metrics} and Table~\ref{table:lstm_optimization:metrics_statistics} suggest that the RCNN performs best when it is optimized with a loss function explicitly designed to model smooth and sharp details found in time-varying signals. In this work, the RMSE and GDL were used to promote such behavior, allowing the modeling of smooth and sharp (i.e.~signal peaks) details attributed to force/torque signals. Nonetheless, other distance functions could potentially be applied for the same purpose. Moreover, these results show that it is important to provide the RCNN with both video sequences and tool data during the training and inference stages.

\begin{figure*}[!t]
	\centering	
	\subfloat[Pushing Task]{\includegraphics[width=1.0\columnwidth]{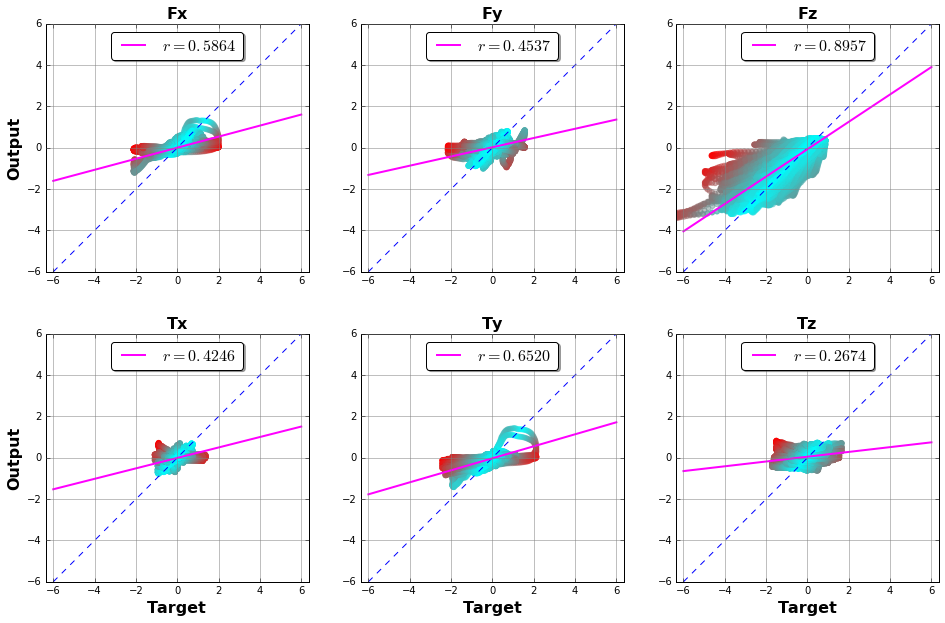}%
	\label{fig:regression_ouput_vs_target:case_III_A:pushing}}
	\hfil
	\subfloat[Pulling Task]{\includegraphics[width=1.0\columnwidth]{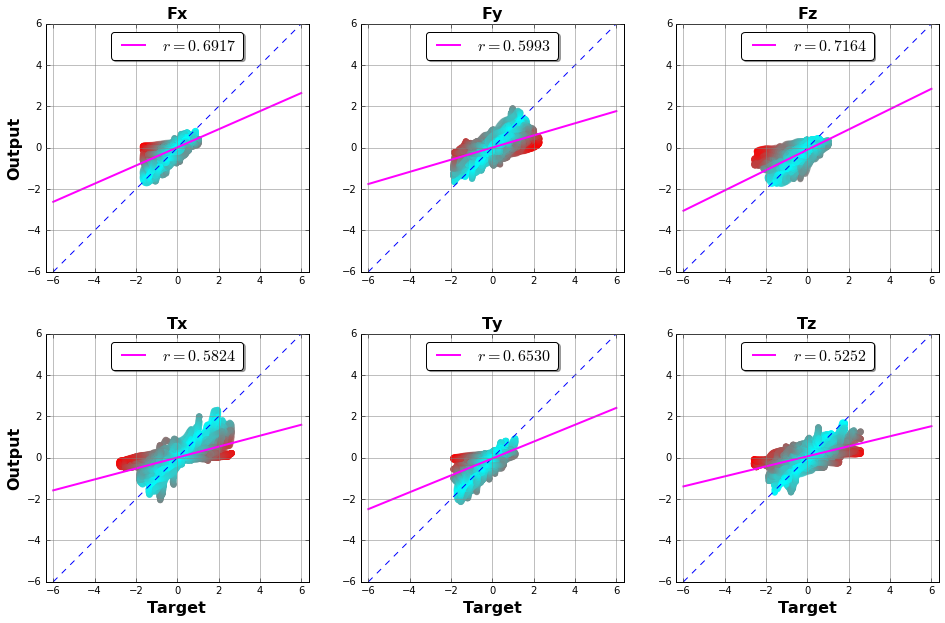}%
	\label{fig:regression_ouput_vs_target:case_III_A:pulling}}
	
	\caption{Case III-A: Neural network output vs target plot (for all data in the test set) related to pushing (left column) and pulling tasks (right column). The Pearson Correlation Coefficient (PCC) is shown for each force component as $r$. The best line that fits the data is shown in magenta color. A perfect fitting to the data is represented by the dotted line in dark blue color. Data points with low and high error are plotted in blue and red colors, respectively.}
	\label{fig:regression_ouput_vs_target:case_III_A}
\end{figure*}

The force estimation quality (from the test dataset) corresponding to case III-A (with the highest accuracy) is described in Fig.~\ref{fig:regression_ouput_vs_target:case_III_A} and Table~\ref{table:force_estimation_quality_in_forcetorque_units}. The neural network output vs target plot and the PCC are shown in Fig.~\ref{fig:regression_ouput_vs_target:case_III_A}, while the RMSE in force and torque units is reported in Table~\ref{table:force_estimation_quality_in_forcetorque_units}. 

In Figs.~\ref{fig:signal_quality_metrics} and~\ref{fig:regression_ouput_vs_target:case_III_A} is observed a high PCC value (0.8957) and low error present in the $F_z$ force component related to pushing tasks. Regarding pulling tasks, the estimated force $F_z$ has also higher PCC value (0.7164) with respect to the rest of force components. However, it falls below the PCC value reported for pushing tasks. These results suggest that interaction forces produced by pushing tasks (smooth signals) are easier to model than those generated by pulling tasks (irregular signals). 
A possible explanation of these results can be deduced from the video frames computed in the space-time domain, depicted in Fig.~\ref{fig:sample_of_preprocessed_video_frames}. 
Thus, when dealing with pushing tasks, tool-tissue interactions seem to be regular and independent of the organs' geometry. For instance, the point of interaction is defined by a small contact area with an oval shape (Fig.~\ref{fig:sample_of_preprocessed_video_frames:pushing}). In contrast, those interactions resulting from pulling tasks are more irregular and highly dependent on the organs' geometry (Fig.~\ref{fig:sample_of_preprocessed_video_frames:pulling}). The slightly imbalance in the dataset samples that represent each surgical task, may be a small contributing factor for this result (59\% and 41\% of the dataset samples correspond pushing and pulling tasks, respectively, as shown in Table~\ref{table:dataset_composition}). 

The results of Table~\ref{table:force_estimation_quality_in_forcetorque_units} show the potential of the proposed RCNN architecture, upon which new models can be devised. For real operational purposes, the RMSE for forces is reported to fall below 0.1 N in both vision-based~\cite{VBFS:Aviles_3_2016} and prototyped sensors~\cite{ForceSensor_SurgForceps:Kim_2015}.

\begin{figure*}[!t]
	\centering	
	\subfloat[Pushing Task]{\includegraphics[width=1.0\columnwidth]{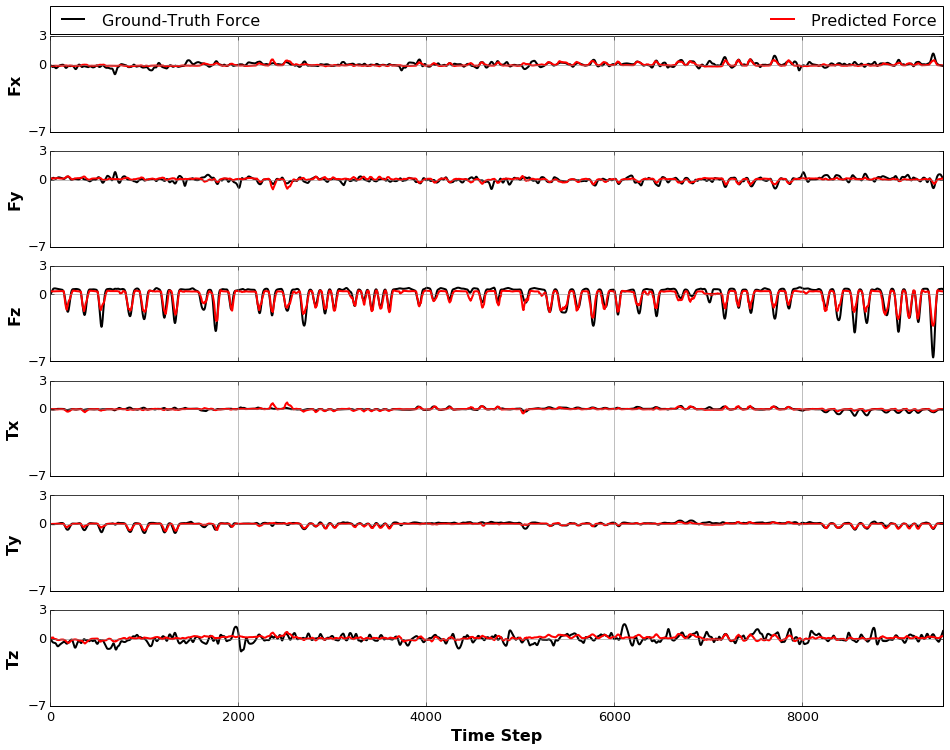}%
		\label{fig:predicted_force:pushing_task}}
	\hfil
	\subfloat[Pulling Task]{\includegraphics[width=1.0\columnwidth]{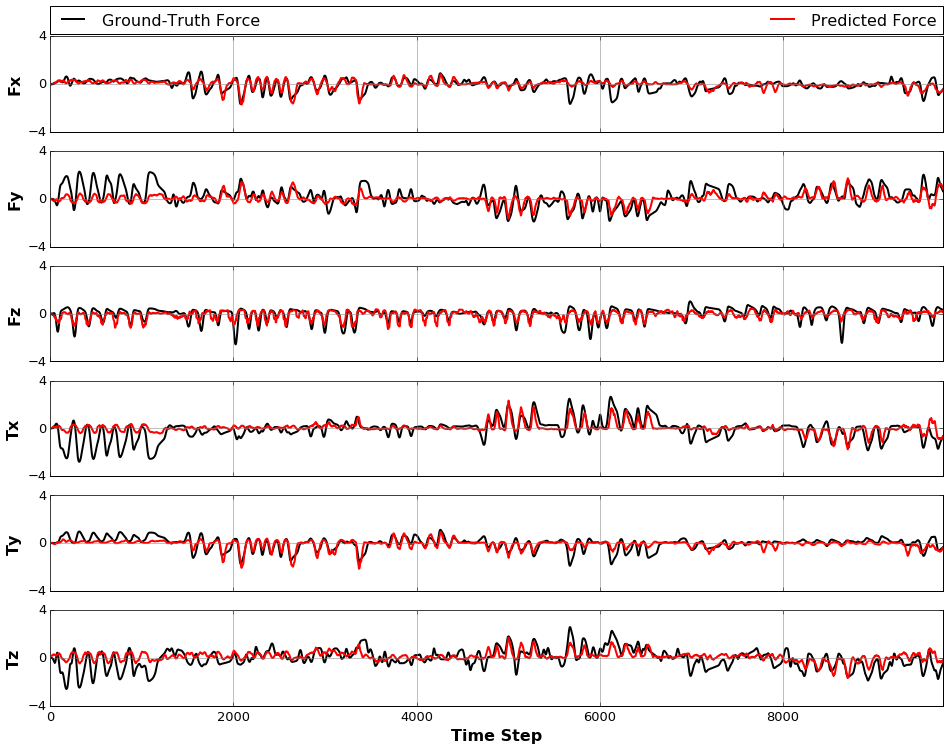}%
		\label{fig:predicted_force:pulling_task}}
	\caption{Case III-A: Sample of estimated interaction forces between tool and tissue (normalized in the range +/-5) for pushing (left column) and pulling tasks (right column).}
	\label{fig:predicted_force}
\end{figure*}

A sample of estimated forces (from the test dataset) between the surgical instrument and the tissue (normalized in the range +/-5), related to case III-A  is shown in Fig.~\ref{fig:predicted_force:pushing_task} and Fig.~\ref{fig:predicted_force:pulling_task} for pushing and pulling tasks, respectively.  Fig.~\ref{fig:predicted_force:pushing_task} shows that the amplitude of most interaction forces (estimated for pushing tasks) are close to zero, with the exception of the $F_z$ force component. The reason is that the forces are mainly applied along the surgical instrument shaft which is aligned with the $z$ axis of the force sensor. It is also observed that the estimated shape of $F_z$ is fully retrieved, although its amplitude differs in some locations from the ground-truth signal. By contrast, in Fig.~\ref{fig:predicted_force:pulling_task} the force and torque components (estimated for pulling tasks) are non-zero, because of the reaction forces applied to the surgical instrument when it is grasping a tissue. Nonetheless, these signals are more difficult to learn in both amplitude and shape.

\subsection{Robustness of the RCNN Model}
\label{subsection:results:rcnn_model_robustness}
\begin{figure*}[!t]
	\centering
	\subfloat[Pushing Tasks]{\includegraphics[width=0.75\columnwidth]{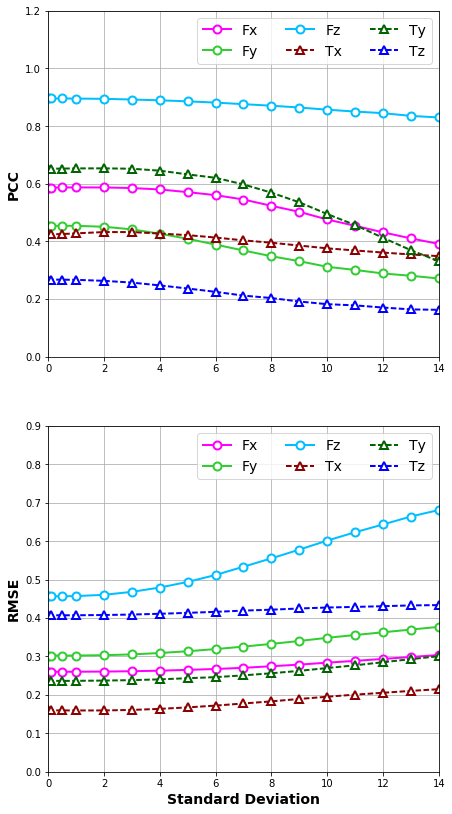}%
	\label{subfig:robustness_against_noise:pushing}}
	\hfil
	\subfloat[Pulling Tasks]{\includegraphics[width=0.75\columnwidth]{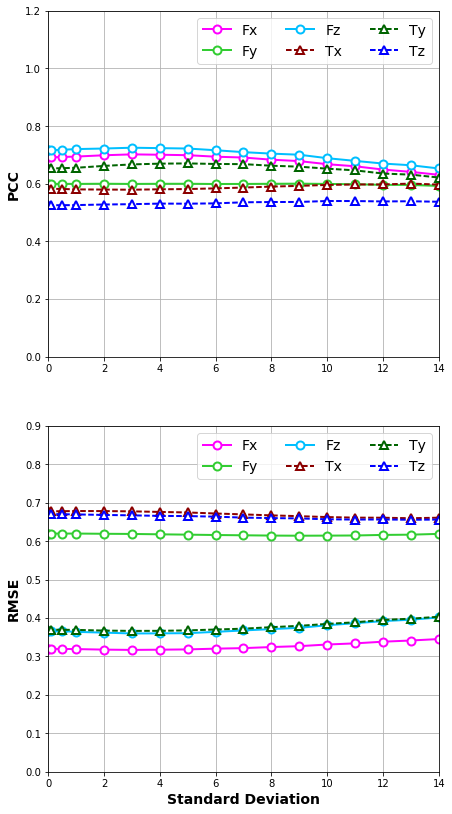}%
	\label{subfig:robustness_against_noise:pulling}}
	\caption{Case III-A: Deterioration of the RCNN model as Gaussian noise is added to tool data with increased strength (by varying the standard deviation). The Pearson Correlation Coefficient (PCC) and Root Mean Squared Error (RMSE) metrics (per force component) related to pushing and pulling tasks are shown on the left and right columns, respectively.}
	\label{fig:robustness_against_noise}	
\end{figure*}

The results of the robustness of the RCNN model against Gaussian noise added on tool data are shown in Fig.~\ref{fig:robustness_against_noise}. In this illustration, it can be observed that the PCC and RMSE metrics are deteriorated by a small margin as the noise intensity is strengthened. Nonetheless, this effect is more noticeable in the metrics related to pushing tasks than those of pulling. These results suggest that the RCNN model is able to cope with tool data corrupted with (Gaussian) noise. Furthermore, they reveal that the estimation of interaction forces heavily relies on the input video sequences. 

The comparison of the RCNN performance by pre-processing video sequences in offline and real-time modes is summarized in Table~\ref{table:rcnn_model_frames_preprocessed_online_vs_offline}. The metrics reported in this table correspond to a pair of video sequences in the test set. Each video sequence is related to pushing and pulling tasks. The relative error reported in this table (shown as a percentage), reveals that the performance of the RCNN model is slightly degraded. A positive relative error represents a deterioration in the quality of the metrics in real-time mode with respect to those in offline mode. The reverse effect, attributed to negative relative errors, is observed in some force components whose contribution is nonsignificant for the task being performed (pushing or pulling).

\subsection{RCNN Model vs ARMAX Model}
\label{subsection:results:rcnn_vs_armax}
The ARMAX and RCNN models are contrasted in Table~\ref{table:rcnn_vs_armax_model}. This table shows the PCC computed from the estimated force signals (data in the test set) by the RCNN (case III-A) and the ARMAX models, related to each surgical task (pushing and pulling). The PCC values presented in this table reveal that the RCNN model is a better choice than the ARMAX model for the task of force estimation.

%
%


\begin{table}[!t] 
	\renewcommand{\arraystretch}{1.15} 
	\caption{Maximum, minimum, and mean values of the Pearson Correlation Coefficient (PCC) and Root Mean Squared Error (RMSE) metrics (shown in Fig.~\ref{fig:signal_quality_metrics}) for all the cases studied (I-A, I-B, ..., III-B). } 
	\label{table:lstm_optimization:metrics_statistics}
	\centering
	\scriptsize
	\begin{tabular}{|l|c @{\hspace{1.5\tabcolsep}}c @{\hspace{1.5\tabcolsep}}c|c @{\hspace{1.5\tabcolsep}}c @{\hspace{1.5\tabcolsep}}c|}
	\hline     
	\bfseries Case 	& \multicolumn{3}{c}{\bfseries Pushing Task} & \multicolumn{3}{c|}{\bfseries Pulling Task} \\ 
	\cline{2-7} 
					& \bfseries Max & \bfseries Min & \bfseries Mean & \bfseries Max & \bfseries Min & \bfseries Mean \\
	\hline
	        		& \multicolumn{6}{c|}{PCC (Values closer to 1.0 are better)}         \\   
	\hline        
	I-A   			& 0.3800           & -0.1351          & 0.0450           & 0.2110           & -0.1732          & 0.0636 \\
	I-B   			& 0.3655           & 0.0406           & 0.1263           & 0.4901           & -0.0241          & 0.2232 \\
	II-A  			& 0.8877           & 0.2474           & 0.5175           & 0.7002           & \bfseries 0.5492 & 0.6100 \\
	II-B  			& 0.8869           & 0.2405           & 0.5097           & 0.7086           & 0.5342           & 0.6024 \\
	III-A 			& \bfseries 0.8957 & \bfseries 0.2674 & \bfseries 0.5466 & \bfseries 0.7164 & 0.5252           & \bfseries 0.6280 \\
	III-B 			& 0.8469           & 0.1841           & 0.4016           & 0.6860           & 0.5367           & 0.6141 \\
	\hline
					& \multicolumn{6}{c|}{ RMSE (Values closer to 0.0 are better)}  \\
	\hline
	I-A   			& 1.1997           & 0.3502           & 0.6407           & 0.8517           & 0.4329           & 0.6509 \\
	I-B   			& 1.3149           & 0.2785           & 0.5672           & 0.8278           & 0.4349           & 0.6313 \\
	II-A  			& \bfseries 0.4531 & 0.1732           & 0.3137           & 0.7043           & 0.3321           & 0.5195 \\
	II-B  			& \bfseries 0.4531 & 0.1726           & 0.3098           & 0.6962           & 0.3419           & 0.5161 \\
	III-A 			& 0.4567           & \bfseries 0.1598 & \bfseries 0.3038 & 0.6778           & \bfseries 0.3199 & \bfseries 0.5041 \\
	III-B 			& 0.6592           & 0.2596           & 0.3967           & \bfseries 0.6756 & 0.3320           & 0.5168 \\
	\hline
\end{tabular}

\end{table}

\begin{table}[!t] 
	\renewcommand{\arraystretch}{1.30} 
	\caption{Case III-A: Root Mean Squared Error (RMSE), where the force and torque units are expressed in Newtons (N) and Newtons per meter (Nm), respectively.} 
	\label{table:force_estimation_quality_in_forcetorque_units}
	\centering
	\scriptsize
	\begin{tabular}{|l|cccccc|}
	\hline
	\bfseries Task 	& $\boldsymbol{F_x}$  	& $\boldsymbol{F_y}$ 	& $\boldsymbol{F_z}$ 	& $\boldsymbol{T_x}$	& $\boldsymbol{T_y}$	& $\boldsymbol{T_z}$ \\
	\hline	
	Pushing 		& 0.1230 				& 0.0892 				& 1.1071 				& 0.2810 				& 0.3621 				& 0.0232 \\	
	Pulling 		& 0.1511 				& 0.1829 				& 0.8894 				& 1.1915 				& 0.5660 				& 0.0381 \\
	\hline
\end{tabular}
\end{table}


\begin{table}[!t] 
	\renewcommand{\arraystretch}{1.150} 
	\caption{Comparison of the performance of the RCNN model in offline (O) and real-time (RT) modes, using Pearson Correlation Coefficient (PCC) and Root Mean Squared Error (RMSE). } 
	\label{table:rcnn_model_frames_preprocessed_online_vs_offline}
	\centering
	\scriptsize
	\begin{tabular}{|l|c @{\hspace{1.5\tabcolsep}} c @{\hspace{1.5\tabcolsep}} c @{\hspace{1.5\tabcolsep}} c @{\hspace{1.5\tabcolsep}} c @{\hspace{1.5\tabcolsep}} c|}
\hline
\bf Mode            & $\boldsymbol{F_x}$  	& $\boldsymbol{F_y}$ 	& $\boldsymbol{F_z}$ 	& $\boldsymbol{T_x}$ 	& $\boldsymbol{T_y}$ 	& $\boldsymbol{T_z}$ \\   
\hline
\multicolumn{7}{|c|}{\bfseries Pushing Task} \\
\hline
                & \multicolumn{6}{c|}{PCC} \\
\cline{2-7}
O	            & 0.5816 				& \bfseries 0.4869		& \bfseries 0.9286 		& \bfseries 0.5860 		& \bfseries 0.8643 		& 0.2432 \\ 
RT	 			& \bfseries 0.5873		& 0.4546 				& 0.8794 				& 0.5480 				& 0.8205 				& \bfseries 0.2611 \\ 
Error~\textsuperscript{$\dagger$} & -0.99 				& 6.64					& 5.29					& 6.49  				& 5.06					& -7.34 \\

\cline{2-7}
				& \multicolumn{6}{c|}{RMSE} \\
\cline{2-7}
O 	 			& \bfseries 0.1797		& \bfseries 0.2182		& \bfseries 0.4528 		& \bfseries 0.1103 		& \bfseries 0.1113 		& 0.3874  \\
RT				& 0.1817 				& 0.2209 				& 0.5918 				& 0.1164 				& 0.1260 				& \bfseries 0.3864  \\
Error~\textsuperscript{$\dagger$} 				& 1.14					& 1.22					& 30.69					& 5.56					& 13.27					& -0.26 \\
\hline
\multicolumn{7}{|c|}{\bfseries Pulling Task} \\
\hline
				& \multicolumn{6}{c|}{PCC} \\
\cline{2-7}
O   	     	& \bfseries 0.7134 		& 0.6635 				& \bfseries 0.7070 		& \bfseries 0.6700 		& 0.7214 				& \bfseries 0.5935 \\		
RT	    		& 0.6838 				& \bfseries 0.6845 		& 0.6547 				& 0.6654 				& \bfseries 0.7238 		& 0.5637 \\
Error~\textsuperscript{$\dagger$} 				& 4.14 					& -3.16					& 7.40					& 0.69					& -0.34					& 5.03 \\
\cline{2-7}
				& \multicolumn{6}{c|}{RMSE} \\
\cline{2-7}
O        		& \bfseries 0.3079  	& 0.5915 				& \bfseries 0.3737 		& 0.6435 				& \bfseries 0.3423		& \bfseries 0.6555 \\	
RT	    		& 0.3217 				& \bfseries 0.5814 		& 0.4009 				& \bfseries 0.6431		& 0.3489 				& 0.6691 \\
Error~\textsuperscript{$\dagger$} 				& 4.48 					& -1.70					& 7.30					& -0.07					& 1.92					& 2.07 \\
\hline
\multicolumn{7}{l}{O: Offline Mode}\\
\multicolumn{7}{l}{RT: Real-Time Mode}\\
\multicolumn{7}{l}{$\dagger$: Relative error in percentage (\%) computed by taking values}\\
\multicolumn{7}{l}{in offline (O) mode as a reference: $Rel.~Error = (RT/O)\times 100~\%$.}
\end{tabular}
\end{table}
\begin{table}[!t] 
	\renewcommand{\arraystretch}{1.150} 
	\caption{Comparison of the Pearson Correlation Coefficient (PCC) computed from the estimated force by the RCNN (case III-A) vs ARMAX models.} 
	\label{table:rcnn_vs_armax_model}
	\centering
	\scriptsize
	\begin{tabular}{|l|c @{\hspace{1.5\tabcolsep}} c @{\hspace{1.5\tabcolsep}} c @{\hspace{1.5\tabcolsep}} c @{\hspace{1.5\tabcolsep}} c @{\hspace{1.5\tabcolsep}} c|}
\hline
\bf Model & $\boldsymbol{F_x}$  	& $\boldsymbol{F_y}$ 	& $\boldsymbol{F_z}$ 	& $\boldsymbol{T_x}$ 	& $\boldsymbol{T_y}$ 	& $\boldsymbol{T_z}$  \\
\hline
\multicolumn{7}{|c|}{\bfseries Pushing Task (PCC)} \\
\hline
RCNN         	& \bfseries 0.5864 		& \bfseries 0.4537  	& \bfseries 0.8957 		& \bfseries 0.4246 		& \bfseries 0.6520 		& \bfseries 0.2674 \\		
ARMAX 		    & 0.0949 				& 0.0166 				& 0.0925 				& 0.0378 				& 0.1331 				& 0.1312 \\
\hline
\multicolumn{7}{|c|}{\bfseries Pulling Task (PCC)} \\
\hline
RCNN 		  	& \bfseries 0.6917 		& \bfseries 0.5993 		& \bfseries 0.7164 		& \bfseries 0.5824 		& \bfseries 0.6530 		& \bfseries 0.5252 \\	
ARMAX 		    & -0.0639 				& 0.2872 				& 0.0593 				& 0.1567 				& 0.0178 				& 0.1709 \\
\hline
\end{tabular}

\end{table}


%
%

\subsection{Requirements for Real Applications}
\label{subsection:results:challenges_real_scenarios}
For practical applications, there are three key features of the RCNN model that should be improved. First, the error reported in Table~\ref{table:force_estimation_quality_in_forcetorque_units}, can be reduced (to meet the design requirement of 0.1 N for forces) by taking into account the processing of depth information. This information can help to improve the quality in the force estimates, similarly in that the addition of tool data (i.e.~the tool-tip trajectory and its grasping status) helped to render force estimates with better quality than processing only video sequences. For this purpose, a monocular depth estimation technique, such as~\cite{Godard_2017:UnsupMonocularDepthEst}, can be used. Second, techniques for pre-processing of video sequences were explored as a first approach to highlight motion due to tool-tissue interactions and ease the learning process of the neural network model. However, an attention model, such as the one described in~\cite{Xu_2015}, represents a suitable approach to automatically learn those image regions that are relevant to the task of interest (force estimation). Finally, the RCNN, consisting of the VGG16 network connected in series with the LSTM-CIFG network, results in a model with many parameters, which is slow during both training and inference stages. For real-time scenarios, a compact model is needed, capable of rendering force estimates without loosing quality. To this end, techniques for compressing and accelerating deep neural networks can be useful. For instance, parameter pruning and sharing, low-rank factorization, transferred/compact convolutional filters, and knowledge distillation~\cite{Cheng_2018:Comp_And_Acc_for_DNNs}.

%
%
%
%
\section{Conclusions \& Future Work}
\label{section:conclusion}
A Recurrent Convolutional Neural Network (RCNN) for Vision-Based Force Sensing (VBFS) in robotic surgery has been developed. The proposed neural network was designed to estimate forces from monocular video sequences, as opposed to the majority of reported works, which rely on stereo vision. For this purpose, a pre-trained CNN was used to learn a compact feature vector representation for each frame in a video sequence ($\phi_t^{video}$), which encodes complex phenomena such as deformation of soft-tissues and specular reflections. This representation together with the tool data ($\phi_t^{tool}$), defined a new feature vector space (i.e.~by concatenating $\phi_t^{video}$ and $\phi_t^{tool}$), increasing the quality in the force estimates. To enforce a temporal constraint, the feature vector space was modeled by an LSTM network. The proposed RCNN model represents an alternative to existing approaches and has the potential to achieve better results in the future. 

From this research work, several experimental findings can be highlighted. First, the force estimation task is achieved better when the CNN and LSTM networks are optimized with a loss function that takes into account the Root Mean Squared Error (RMSE) and Gradient Difference Loss (GDL). The intuition behind this loss function design is that continuous and time-varying signals can be interpreted as composed of smooth and sharp details. Therefore, the RMSE addresses the modeling of smooth information found in force/torque signals (i.e.~sine wave-like signals), while the GDL promotes the modeling of sharp details attributed to these signals (i.e.~signal peaks). However, other alternatives to the GDL may result in better outcomes. For instance, the adversarial loss, which is derived from the Generative Adversarial Network (GAN) framework~\cite{Goodfellow_2014}, has proven useful in the modeling of high-frequency components found in images. This type of loss can be adapted to the modeling of sharp details found in force/torque signals. Second, both video sequences and tool data, provide important cues for the estimation of forces than using either source of information alone. Third, this study shows that interaction forces resulting from pushing tasks (characterized by smooth signals) are easier to model and estimate than those produced by pulling tasks (characterized by irregular signals). Fourth, the experiment related to the robustness of the RCNN against Gaussian noise added to the tool data suggests that the RCNN model is able to cope with this perturbation. Furthermore, this experiment shows that the RCNN relies heavily on video sequences to estimate interaction forces. Fifth, regarding the  pre-processing of video sequences in real-time, this experiment shows that the RCNN model performance is slightly degraded with respect to that relying on video sequences pre-processed offline. Finally, the ARMAX model is unable to model the complex relationship between tool data and interaction forces. Therefore, the RCNN model represents a better choice in the task of force estimation. 

The RCNN model presented in this work addresses a special case of real surgical scenarios. The camera and organs are static while the surgical instrument is in motion. The proposed RCNN model has been evaluated only in static scenarios, using a dataset enriched with video sequences recorded from different viewpoints. This allows the neural network to learn the relation between tool-tissue interactions and force under a variety of perspectives. A real scenario is usually more dynamic, with the camera moving automatically or at surgeon's will. Moreover, the organs may be affected by physiological motion due to breathing and heart beating cycles. 

As future work, five research directions can be explored. Some of them have already been discussed in Section~\ref{subsection:results:challenges_real_scenarios}. First, for real operational purposes, the force estimation quality, shown in Table~\ref{table:force_estimation_quality_in_forcetorque_units}, could be improved by taking into account depth information (i.e.~using a technique, such as~\cite{Godard_2017:UnsupMonocularDepthEst}). Second, a model designed in a semi-supervised learning setting using an Auto-Encoder network and GANs, represents a potential approach to find a suitable feature vector representation from video sequences when few data are available. Third, incorporating an attention model~\cite{Xu_2015}, would allow automatically select those regions in video sequences that contribute to the learning process (i.e.~where tool-tissue interactions are present), avoiding the need of applying pre-processing operations (i.e.~mean frame removal and space-time transformation). Moreover, this attention mechanism would allow the extension of the neural network model to the estimation of forces related to more complex surgical tasks than pushing and pulling (i.e.~suturing or knot-tying) and its application to dynamic scenarios (i.e.~by processing motion due to uniquely tool-tissue interactions, while suppressing the motion caused by the camera and organs). Fourth, techniques for compressing and accelerating deep neural networks should be investigated. They will help in designing a compact neural network model suitable for real-time scenarios. Finally, a better understanding of the RCNN model, e.g., by interpretation of its predictions \cite{BachPLOS15, MonDSP18}, will certainly help in designing more efficient RCNN architectures in the future.

%
%
\section*{Acknowledgment}
The first author of this work acknowledges the Mexican National Council for Science and Technology (CONACYT) and the Mexican Secretariat of Public Education (SEP) for their support in doctoral studies. The work is supported by the Ministerio de Econom\'{i}a y Competitividad and the Fondo Europeo de Desarrollo Regional, ref. DPI2015-70415-C2-1-R (MINECO/FEDER).

%
%

\section*{References}
\bibliography{References/Reduced_Version/Introduction,References/Reduced_Version/RelatedWorks,References/Reduced_Version/KeyPapers}

\end{document}